\documentclass[11pt]{article}
\usepackage[final]{acl}

\usepackage{times}
\usepackage{latexsym}
\usepackage[T1]{fontenc}
\usepackage[utf8]{inputenc}
\usepackage{microtype}
\usepackage{inconsolata}
\usepackage{graphicx}
\usepackage{booktabs}
\usepackage{multirow}
\usepackage{array}
\usepackage{amsmath}
\usepackage{placeins}
\usepackage{pgfplots}
\pgfplotsset{compat=1.18}
\usepgfplotslibrary{groupplots}
\renewcommand{\thefootnote}{\fnsymbol{footnote}}
\usepackage{xurl}
\title{Instruction-Tuned Language Models Cannot Sample \\ from Distributions They Can Describe}

\author{
Chaemin Jang$^{1}$, Jihee Kim$^{1,2,3}$\footnotemark[2], Dongman Lee$^{1}$\footnotemark[2] \\
$^{1}$School of Computing, KAIST \\
$^{2}$School of Business and Technology Management, KAIST \\
$^{3}$Graduate School of Data Science, KAIST \\
\texttt{jchaemin@kaist.ac.kr}
}

\begin{document}
\maketitle
\renewcommand{\thefootnote}{\fnsymbol{footnote}}
\footnotetext[2]{Co-corresponding authors.}
\renewcommand{\thefootnote}{\arabic{footnote}}

\begin{abstract}
Silicon sampling uses language models as proxies for human survey respondents, treating each model call as an independent draw from the persona's response distribution. We show this draw does not exist: instruction-tuned models do not sample from distributions, they collapse to a single output. 
The same persona on the same question returns the same answer on more than half of items in a public-opinion benchmark, and the model's internal probabilities concentrate on a single option. 
The failure is associated with, and amplified by, instruction-targeted post-training: instruction-tuned models are worse than their own bases in every family we can compare, the gap widens at each successive post-training stage and with the size of the tuning update, and continued pretraining on non-instruction tokens leaves it unchanged. 
Yet the knowledge survives: the same model that cannot sample from a distribution can describe it accurately in a single call. We call this gap the KNOWS/DOES split. Exploiting the split, a single call that asks the model to describe the response distribution more than halves the error against human survey data compared to persona aggregation. When per-persona outputs are required, we propose Prompt-Perturbed Argyle (PPA), which reduces the same error by 21\%, spreading each persona's answers to mirror real population differences at no added cost. 
\end{abstract}

\begin{figure*}[t!]
\centering
\includegraphics[width=0.55\textwidth, angle=-90]{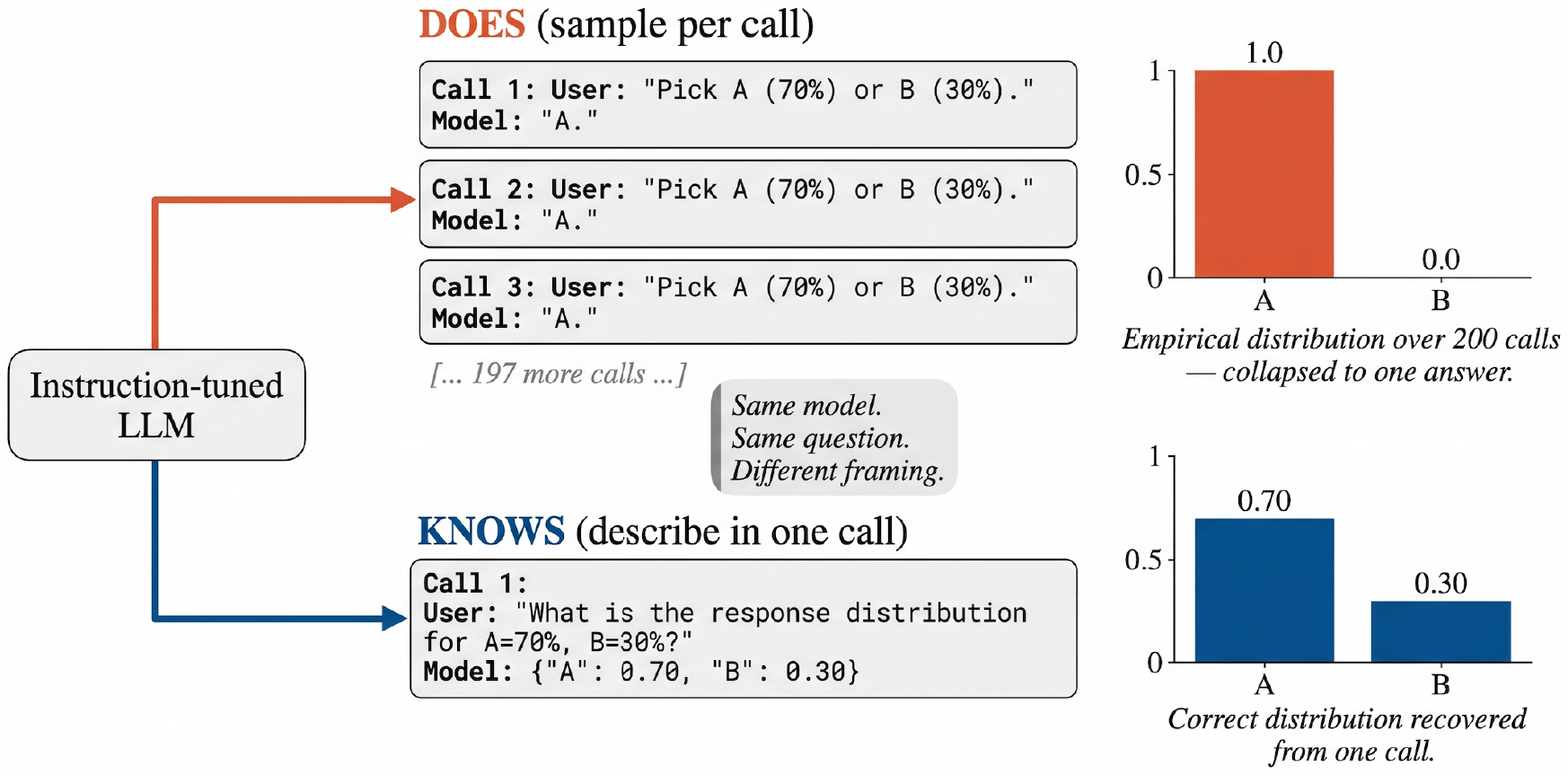}
\caption{The KNOWS/DOES split on the skewed binary target. When asked to sample one outcome per call, the instruction-tuned model collapses to a single answer across 200 repeated calls (empirical: A=1.00, B=0.00). When asked to describe the distribution in one call, the same model recovers the correct shape (A=0.70, B=0.30). The pattern holds across all five tasks tested (\S\ref{sec:failure-knowsdoes}; Table~\ref{tab:knows-does}). Failure persists across temperature, top-$p$, and prompt-level corrections.}
\label{fig:knowsdoes}
\end{figure*}
\section{Introduction}
\label{sec:intro}

Silicon sampling uses language models as proxies for human survey respondents \citep{argyle2023out, park2023generative, santurkar2023whose}. The standard pipeline prompts the model with a set of personas, treats each call as one draw from the corresponding persona-conditional response distribution, and aggregates the responses to produce a population estimate. The method has been adopted across agent-based simulation \citep{park2023generative, park2024generative} and survey methodology \citep{horton2023large, aher2023using}. A growing literature reports that silicon sampling pipelines systematically miscalibrate the human distributions they target \citep{bisbee2024synthetic, boelaert2024machine, tjuatja2024llms, hu2024persona}, usually attributed to training-data gaps or alignment bias. We argue the cause is more basic: the per-call output of an instruction-tuned language model is not a sample from the conditional distribution the pipeline assumes.

We characterize this failure on a controlled set of synthetic categorical targets. Asked for a uniform random integer in 1--100, an instruction-tuned model returns 42 in 78\% of calls (Table~\ref{tab:rng-primitive}). Across seven categorical targets covering skewed, bimodal, and multi-way distributions, the model produces a single answer in over 94\% of calls. The failure is visible at the representational level: the top-two logit gap at temperature zero exceeds 14 nats on some targets (\S\ref{sec:failure-categorical}). Algorithmic chain-of-thought scaffolds do not address the failure either, because the random number the model draws inside the scaffold is itself non-uniform.

The KNOWS/DOES split (Figure~\ref{fig:knowsdoes}) has direct consequences for silicon sampling. On 100 OpinionQA items \citep{santurkar2023whose} drawn from Pew American Trends Panel surveys \citep{pew_atp}, standard persona aggregation fails to produce variation: across 50 repeated calls on the same (persona, item) pair, 57\% of pairs return the same answer every time, and 80\% return the same answer in over 90\% of calls. The describe pathway recovers what aggregation cannot. A single call asking the model for the response distribution directly cuts the gap to the human results by more than half (TV distance 0.22 vs.\ 0.46), and the advantage holds even when personas are drawn to match the actual Pew respondents on demographics. Some applications still need a separate output per simulated respondent. For those, we propose Prompt-Perturbed Argyle (PPA), which randomizes option ordering and question phrasing across calls. PPA reduces the same gap by 21\% at no added cost, driven by the per-call answer variation it induces.

\paragraph{Contributions.} Prior work has documented mode collapse under RLHF \citep{padmakumar2024diversity}, silicon sampling miscalibration \citep{bisbee2024synthetic, boelaert2024machine, santurkar2023whose}, and the asymmetry that describing a distribution outperforms aggregating per-call samples \citep{meister2025benchmarking, xiao2025knowledge, zhang2025verbalized}. We make four contributions.\footnote{\label{fn:repo}\url{github.com/jchaemin/instruct-sampling-collapse}}
\begin{enumerate}
\item A controlled characterization of the per-call sampling failure, including a chain-of-thought analysis showing why algorithmic scaffolding cannot fix it (\S\ref{sec:failure-knowsdoes}).
\item Evidence that the failure is associated with, and amplified by instruction-targeted post-training, across three model families with different pipelines, grows monotonically across post-training stages, with the size of the instruction-tuning update, and is absent under continued pretraining (\S\ref{sec:failure-alignment}).
\item A demographically-matched Argyle baseline that closes the persona-mismatch confound in prior describe-vs-sample evaluations (\S\ref{sec:silicon}).
\item Prompt-Perturbed Argyle (PPA): a new method for per-respondent applications that reduces TV by 21\% at no added cost, with a controlled mechanism analysis (\S\ref{sec:ppa}).
\end{enumerate}

\section{Related Work}
\label{sec:related}

\paragraph{Silicon sampling and its miscalibration.}
LLMs-as-respondents originated with \citet{argyle2023out} and the paradigm has been extended to opinion dynamics \citep{park2023generative, park2024generative}, social experiments \citep{aher2023using}, and economic decisions \citep{horton2023large}. A parallel literature documents systematic miscalibration of these pipelines: LLM-generated populations show smaller effective sample sizes than the human populations they approximate \citep{bisbee2024synthetic}, between-group variance collapses \citep{boelaert2024machine}, the resulting estimates skew toward specific demographic viewpoints \citep{santurkar2023whose}, response biases diverge from human survey patterns \citep{tjuatja2024llms}, and prompt-framing sensitivity destabilizes them further \citep{hu2024persona}. \citet{dominguezolmedo2024questioning} show that re-ordering options alone shifts the answer distribution toward uniform; we ask whether such perturbation, pooled across calls, recovers the target rather than merely flattening it. These failures are typically attributed to representational gaps in training data or to social biases acquired through alignment. We identify a more proximate cause: the per-call sampling primitive is structurally degenerate in instruction-tuned models, and this failure propagates into every aggregate the pipeline produces.

\paragraph{Verbalized sampling and describer--executor asymmetry.}
\citet{meister2025benchmarking} document that asking the model to describe a distribution outperforms aggregating per-call samples on opinion benchmarks, \citet{xiao2025knowledge} show the same asymmetry on coin flips, and \citet{zhang2025verbalized} propose Verbalized Sampling (VS), which prompts the model to verbalize the distribution as JSON probabilities. The describe pathway we evaluate is structurally identical to VS, and on our 100-item OpinionQA benchmark the two are statistically indistinguishable (Wilcoxon $p = 0.14$). We extend this line of work mechanistically, tracing the asymmetry to a degenerate sampling primitive visible in the logits and linking it to instruction-targeted post-training, and methodologically, with PPA for pipelines where single-call description cannot substitute.

\paragraph{Mode collapse, RLHF, and decoding.}
Instruction-tuned and RLHF-aligned models exhibit reduced output diversity relative to base models \citep{padmakumar2024diversity}, with proposed mechanisms including reward-model overfitting \citep{ouyang2022training, stiennon2020learning} and regularization toward a peaked reference policy \citep{glaese2022improving}. Standard decoding controls (temperature, top-$k$, top-$p$) operate on a next-token distribution that is fixed at inference time \citep{holtzman2019curious, fan2018hierarchical}. We sharpen this picture in two ways. First, the collapse on uniform-target prompts is categorical rather than statistical: more than 94\% of calls produce the same option, and the logit gap at $T=0$ exceeds 14 nats on some targets, far beyond any temperature the API permits. Second, our base-vs-instruct comparison establishes the alignment-induced direction empirically.

\paragraph{Formal versus functional competence.}
The KNOWS/DOES split sits within the broader dissociation \citet{mahowald2024dissociating} call formal vs.\ functional competence: models can introspect on knowledge better than they can produce it \citep{kadavath2022language}, can be elicited to express calibrated uncertainty over outputs they cannot generate correctly \citep{lin2022teaching}, and exhibit verbalize--execute splits on procedural and probabilistic tasks \citep{xiao2025knowledge}. We extend the frame to distributional generation, providing a specific mechanism in the form of a degenerate sampling primitive visible at the logit level and tracing it to instruction-targeted post-training.

\paragraph{LLMs as data sources and in-context learning.}
The use of LLMs for synthetic training data \citep{wang2022self, honovich2022unnatural}, evaluation benchmarks \citep{liu2022wanli}, and information retrieval queries \citep{bonifacio2022inpars} shares silicon sampling's assumption that per-call outputs are draws from a target. Our findings imply this assumption requires verification for any aggregate-dependent pipeline. The KNOWS pathway's success connects to explanations of in-context learning as structured pattern-matching \citep{brown2020language, wei2022chain, kojima2022large, xie2021explanation, min2022rethinking}. PPA exploits persona conditioning \citep{park2022social, aher2023using} not as a sampling mechanism but as a way to land each call on a different deterministic answer.

\section{Characterizing the Per-Call Sampling Failure}
\label{sec:failure}

This section characterizes the per-call sampling failure on a controlled set of synthetic categorical targets. We document the failure at three levels: how the model samples from a stated target distribution, what its logits look like at temperature zero, and what it does when asked for a random number with no target specified. The third level matters: the natural fix for the first is to walk the model through a sampling algorithm via chain-of-thought. But this depends on the model producing a random number internally, and we show this primitive is broken.

\subsection{Setup}
\label{sec:failure-setup}

Unless noted, we use \texttt{gpt-4o} at temperature 1.0 with 200 calls per condition for behavioral runs, and temperature 0 with top-20 logprobs for logit reads. The base-vs-instruct comparison of \S\ref{sec:failure-alignment} additionally uses \texttt{Llama-3.1-8B}, \texttt{Mistral-7B-v0.3}, and \texttt{Qwen2.5-7B} with their instruction-tuned counterparts. Full setup, prompts, and parsing rules are in Appendix~\ref{app:setup}; cross-family replication on five additional instruction-tuned families is in Appendix~\ref{app:cross-family}. Throughout, we report total-variation (TV) distance between the empirical distribution of 200 calls and the target distribution, which is bounded in $[0, 1]$ with $0$ indicating perfect match.

\subsection{The per-call draw does not exist}
\label{sec:failure-categorical}

The model handles only the degenerate boundary case: a point target (``always output 5'') is reproduced exactly. Every other shape fails, the uniform included (Table~\ref{tab:seven-targets-inline}; full per-target results in Appendix~\ref{app:seven-targets}): across the five structured targets of Table~\ref{tab:seven-targets-inline}, the model produces a single answer in at least 98\% of calls, and on the uniform 1--5 target in 94.3\%.

\begin{table}[h!]
\centering
\small
\setlength{\tabcolsep}{4pt}
\resizebox{\columnwidth}{!}{%
\begin{tabular}{lcc}
\toprule
Target & Target distribution & Empirical distribution \\
\midrule
Skewed binary & $[0.70, 0.30]$ & $[1.00, 0.00]$ \\
Mixture & $[0.70, 0.30]$ & $[1.00, 0.00]$ \\
Three-way & $[0.50, 0.30, 0.20]$ & $[0.98, 0.02, 0.01]$ \\
Bimodal & $[0.5, 0, 0, 0, 0.5]$ & $[1.00, 0, 0, 0, 0]$ \\
Skewed 5-way & $[0.40, 0.30, 0.15, 0.10, 0.05]$ & $[1.00, 0.00, 0.00, 0.00, 0.00]$ \\
\bottomrule
\end{tabular}
}
\caption{Per-target results across the five non-trivial targets, 200 calls each. The empirical distribution concentrates on a single answer in at least 98\% of calls on every target, regardless of the shape the target requests.}
\label{tab:seven-targets-inline}
\end{table}

Two features of Table~\ref{tab:seven-targets-inline} are not mere miscalibration. First, the bimodal row shows the model honoring the zero-probability constraint perfectly (mass 0 on options 2, 3, 4) while completely ignoring the 50/50 part of the same instruction. The same prompt simultaneously satisfies a point-shaped constraint and violates a distribution-shaped one. Second, swap tests confirm that the model is reading the instruction rather than producing a fixed default: swapping a target from $[A{=}0.7, B{=}0.3]$ to $[A{=}0.3, B{=}0.7]$ relocates the model's single output from A to B but never produces the intended distribution.

The failure is not sampling noise either. At temperature zero, the top-two logit gap ranges from 3.75 to over 16 nats across our targets (Appendix~\ref{app:logit}), and recovering a 70/30 target from a 14-nat gap would require a sampling temperature near 17, far above the API maximum of 2. Decoding adjustments cannot fix the failure because it occurs before sampling.

\subsection{The random-number primitive is broken}
\label{sec:failure-rng}
The failure sits before sampling, so we probe where it originates: direct random-number prompts with no persona, distribution, or task framing.

Asked to ``pick a uniform random $X$'' across five output formats, the model concentrates on a format-specific favored value in every case (Table~\ref{tab:rng-primitive}). All five reject uniformity at $p < 0.001$.

\begin{table}[h!]
\centering
\small
\begin{tabular}{lcc}
\toprule
Format & Favored value & Share \\
\midrule
Integer 1--100 & 42 & 78\% \\
Integer 1--10 & 7 & 97\% \\
Alphabet A--Z & G & 47\% \\
Coin flip & heads & 85\% \\
Float $[0,1]$ & 0.55 (std 0.11) & --- \\
\bottomrule
\end{tabular}%

\caption{The model concentrates on a format-specific favored value when asked for a uniform random sample (200 calls per format). All five formats reject uniformity at $p < 0.001$. Full results in Appendix~\ref{app:rng}.}
\label{tab:rng-primitive}
\end{table}

The concentration is not a lexical effect. Excluding the modal value by name relocates the mass to a different favored value: ``Do not pick 42'' yields 37 in 52\% of calls. Cross-family replication shows the same pattern on five other instruction-tuned families (Appendix~\ref{app:cross-family}), each concentrating on a favored value (73, 47, 53, 42, and 1 on the 1--100 task). Algorithmic chain-of-thought scaffolds rely on this same primitive (\S\ref{sec:failure-knowsdoes}).

\section{Instruction-Targeted Post-Training Amplifies the Failure}
\label{sec:failure-alignment}

The failure is not inherent to language modeling. We compare each instruction-tuned model against its pretrained base counterpart on a five-target panel, across three families with materially different post-training pipelines: Llama-3.1-8B (RLHF-based), Mistral-7B-v0.3 (publicly described as supervised fine-tuning without RLHF), and Qwen2.5-7B (a multi-stage pipeline that includes DPO). Full protocol is in Appendix~\ref{app:base-instruct}.

\begin{table}[h]
\centering
\small
\setlength{\tabcolsep}{4pt}
\resizebox{\columnwidth}{!}{%
\begin{tabular}{lcccccc}
\toprule
 & \multicolumn{2}{c}{Llama-3.1-8B} & \multicolumn{2}{c}{Mistral-7B-v0.3} & \multicolumn{2}{c}{Qwen2.5-7B} \\
\cmidrule(lr){2-3} \cmidrule(lr){4-5} \cmidrule(lr){6-7}
Task & Base & Inst & Base & Inst & Base & Inst \\
\midrule
uniform 1--5      & 0.11 & \textbf{0.45} & 0.15 & \textbf{0.40} & 0.18 & \textbf{0.80} \\
fair coin         & 0.05 & \textbf{0.41} & 0.11 & \textbf{0.34} & 0.09 & \textbf{0.50} \\
skewed binary     & 0.15 & 0.02 & 0.03 & \textbf{0.30} & 0.08 & \textbf{0.30} \\
bimodal 5-way     & \textbf{0.33} & \textbf{0.35} & \textbf{0.37} & \textbf{0.73} & \textbf{0.30} & \textbf{0.60} \\
skewed 5-way      & \textbf{0.30} & \textbf{0.45} & \textbf{0.32} & \textbf{0.49} & \textbf{0.36} & \textbf{0.50} \\
\midrule
mean              & 0.19 & 0.34 & 0.20 & 0.45 & 0.20 & 0.54 \\
\# fail ($>$0.20) & 2/5  & 4/5  & 2/5  & 5/5  & 2/5  & 5/5 \\
\bottomrule
\end{tabular}%
}
\caption{V0 (plain sampling) TV-to-target across three families, base vs.\ instruction-tuned counterpart, 200 calls per cell, seed 13. Bold cells indicate failure at the TV $> 0.20$ threshold. Targets: uniform $[0.2]^{5}$, fair coin $[0.5,0.5]$, skewed binary $[0.7,0.3]$, bimodal 5-way $[0.4,0.1,0,0.1,0.4]$, skewed 5-way $[0.5,0.2,0.1,0.1,0.1]$; the two 5-way targets differ from those of Table~\ref{tab:seven-targets-inline}. A second seed reproduces every mean ordering (base/instruct 0.21/0.33, 0.16/0.45, 0.22/0.54).}
\label{tab:base-instruct}
\end{table}

Across all three families, the instruction-tuned model has higher mean TV than its pretrained base counterpart (Table~\ref{tab:base-instruct}), with mean gaps of 0.15 (Llama), 0.26 (Mistral), and 0.34 (Qwen). The base models are not reliable samplers either (every base fails both 5-way targets); instruction tuning amplifies an existing tendency. Mistral-v0.3 is SFT-centered with no RLHF stage, yet the gap still appears, so the effect is not RLHF-specific.

Two further experiments isolate instruction-targeting from additional training in general. Within the OLMo-2 lineage, whose full pipeline is released, TV-to-target grows monotonically at every post-training stage (Table~\ref{tab:staging}a: base $0.236$, SFT $0.321$, DPO $0.411$, RLVR $0.442$), a dose-response rather than a single-stage artifact. Yet continued pretraining alone does not degrade sampling: Qwen2.5-Coder, which adds trillions of non-instruction tokens to the same base, matches it ($0.168$ vs.\ $0.203$) while its instruction-tuned sibling collapses to $0.540$ (Table~\ref{tab:staging}b). Non-instruction training leaves the primitive intact.

The same conclusion follows when the instruction-tuning update is dialed in continuously rather than compared at two endpoints. Writing $\tau = \theta_{\text{tuned}} - \theta_{\text{base}}$ for the weight difference between a base checkpoint and its instruction-tuned successor, we evaluate $\theta_{\text{base}} + \alpha\tau$ for $\alpha \in \{0, 0.25, 0.5, 0.75, 1, 1.25\}$ with the prompt held fixed, so only the weights change. TV-to-target rises with $\alpha$ on both lineages we test --- OLMo-2 base$\to$SFT from $0.236$ at $\alpha = 0$ to $0.317$ at $\alpha = 1$ and $0.410$ at $\alpha = 1.25$; Qwen2.5 base$\to$Instruct from $0.198$ to $0.459$ to $0.519$ --- and the temperature-zero probability of the modal answer rises in step ($0.40 \to 0.55$ and $0.59 \to 0.95$). Removing the update ($\alpha = 0$) restores the base behavior; amplifying it past the released checkpoint ($\alpha = 1.25$) deepens the collapse (Appendix~\ref{app:base-instruct}, Table~\ref{tab:knockout}).

\begin{table}[h]
\centering
\small
\setlength{\tabcolsep}{6pt}
\begin{tabular}{lc}
\toprule
Model / stage & Mean TV-to-target \\
\midrule
\multicolumn{2}{l}{\emph{(a) OLMo-2 post-training stages}} \\
Base (pretrained)       & 0.236 \\
$+\,$SFT                & 0.321 \\
$+\,$DPO                & 0.411 \\
$+\,$RLVR               & 0.442 \\
\midrule
\multicolumn{2}{l}{\emph{(b) Qwen2.5 continued-pretraining control}} \\
Base                    & 0.203 \\
Coder (non-instruction) & \textbf{0.168} \\
Instruct                & \textbf{0.540} \\
\bottomrule
\end{tabular}
\caption{Post-training decomposition. (a) Within the OLMo-2 lineage, TV-to-target grows at every successive post-training stage. (b) Continued pretraining on non-instruction tokens (Coder) does not degrade sampling, whereas instruction tuning on the same base does. Mean TV over the five-target panel of Table~\ref{tab:base-instruct}.}
\label{tab:staging}
\end{table}

\section{The KNOWS/DOES Split}
\label{sec:failure-knowsdoes}

\subsection{The model can describe what it cannot sample}
The failure lives in the sampling primitive; does the distributional knowledge survive in another form? We test this through a controlled ten-intervention study on the five-target panel (Table~\ref{tab:knows-does}). Eight interventions (V0--V7) instruct the model to produce one sample per call, including six prompt-level corrections and algorithmic chain-of-thought. Two interventions (V8, V9) instruct the model to describe the distribution in a single call, either as a list of $N$ exemplars or as JSON probabilities. Full prompts are in the released code.

\begin{table}[h]
\centering
\small
\begin{tabular}{clcc}
\toprule
ID & Intervention & Type & TV \\
\midrule
V0 & Plain sampling & DOES & 0.52 \\
V1 & Verbatim feedback & DOES & 0.47 \\
V2 & Apologetic framing & DOES & 0.48 \\
V3 & Third-person framing & DOES & 0.50 \\
V4 & Explicit bias-naming & DOES & 0.52 \\
V5 & In-context demos & DOES & 0.50 \\
V6 & Step-by-step reasoning & DOES & 0.43 \\
V7 & Algorithmic CoT & DOES & 0.48 \\
\midrule
V8 & List of $N$ exemplars & KNOWS & \textbf{0.03} \\
V9 & JSON probabilities & KNOWS & \textbf{0.03} \\
\bottomrule
\end{tabular}
\caption{Ten prompt-level interventions on five categorical tasks, mean TV across tasks (200 calls per task). Every DOES intervention fails ($\text{mean TV} \geq 0.43$, with 38 of 40 per-task cells above 0.20); both KNOWS interventions succeed within 0.05 TV on every task. Per-task TV and full intervention text in Appendix~\ref{app:knows-does}.}
\label{tab:knows-does}
\end{table}

The pattern in Table~\ref{tab:knows-does} is sharp: every DOES intervention fails, but both KNOWS interventions succeed. We call this gap between describing and sampling the \emph{KNOWS/DOES split}: the model knows what the target distribution looks like but cannot enact that knowledge through per-call output.

\subsection{Why describing is easier than sampling} 
For current instruction-tuned models, describing a distribution in one call is an easier task than producing independent per-call draws from it. We treat this asymmetry as the point rather than a confound. Deployed pipelines depend on the harder path (one collapsed draw per call, pooled); the easier path, a single distributional query, is the accurate one. The knowledge is intact; what fails is emitting one calibrated draw at a time (\S\ref{sec:failure}).

\section{Application to Silicon Sampling on OpinionQA}
\label{sec:silicon}

This section applies the KNOWS/DOES framework to silicon sampling on OpinionQA \citep{santurkar2023whose}, which maps Pew American Trends Panel survey items \citep{pew_atp} to 5-point Likert responses, using the Pew distribution as ground truth.

\subsection{Setup}
\label{sec:silicon-setup}

We use 100 OpinionQA items selected by a pre-registered rule from a fixed wave list, restricted to 5-point Likert items with usable ground truth, with each item run under 100 personas sampled from the joint distribution of six demographic attributes (race, sex, age, education, income, party) in Pew ATP Wave 92 following the Argyle protocol \citep{argyle2023out}. The same persona set is used for all standard-Argyle, PPA, and decoding-sweep experiments, with \texttt{gpt-4o} at temperature 1.0. Full setup is in Appendix~\ref{app:setup}.
We evaluate three pipelines:

\begin{itemize}
\setlength{\itemsep}{2pt}
\item \textbf{Standard Argyle} (baseline): for each (persona, item) pair, we make one call with the persona description, question, and response options, and aggregate the 100 responses per item.

\item \textbf{Demographically-matched Argyle}: for each item, we sample 100 actual respondent rows (with replacement, using the wave's survey weights) from the Pew respondents who answered that item and render each with the same six-attribute persona template as standard Argyle. Otherwise identical to standard Argyle. Full protocol in Appendix~\ref{app:matched-argyle}.

\item \textbf{Describe pathway}: one call per item, no personas. The prompt asks the model to return the population response distribution directly as JSON probabilities. Full protocol in Appendix~\ref{app:matched-argyle}.
\end{itemize}

\subsection{Results}
\label{sec:silicon-results}

We additionally run 50 repeated calls on 100 cells (20 items $\times$ 5 personas) to measure per-(persona, item) Shannon entropy for the persona-aggregation pipelines (Appendix~\ref{app:per-cell-entropy}, Table~\ref{tab:per-cell-entropy}).

\begin{table}[h!]
\centering
\small
\setlength{\tabcolsep}{6pt}
\resizebox{\columnwidth}{!}{%
\begin{tabular}{lccc}
\toprule
Pipeline & Per-call entropy & Deterministic cells & TV-to-Pew \\
\midrule
Standard Argyle     & 9\%  & 57\% & 0.46 \\
Describe pathway    & ---  & ---  & \textbf{0.22} \\
\bottomrule
\end{tabular}%
}
\caption{Standard Argyle vs.\ the describe pathway on 100 OpinionQA items. Per-call entropy is as a fraction of the 5-Likert maximum. Matched Argyle is reported separately (Appendix~\ref{app:matched-argyle}); its TV ($0.484$) is statistically indistinguishable from standard Argyle's, ruling out persona-mismatch as the explanation.}
\label{tab:silicon-results}
\end{table}

The categorical collapse documented in \S\ref{sec:failure} on synthetic targets replicates on real opinion items: most (persona, item) cells return the same answer every call. Demographic matching does not reduce TV (matched Argyle TV $= 0.484$ vs.\ standard $= 0.459$; paired Wilcoxon $p = 0.092$, $n = 100$; Appendix~\ref{app:matched-argyle}), ruling out persona-mismatch as the explanation for persona aggregation's poor performance. The gap is not which personas are sampled but whether each call samples one response (DOES) or describes the population (KNOWS).

\paragraph{Beyond the 5-point Likert format.} On 24 Pew items with 2--4 options (8 each, $\geq 200$ respondents), 100 persona calls per item per pipeline on three open models, describe beats sampling on all three ($0.184$/$0.254$, $p=0.025$; $0.199$/$0.299$, $p=0.006$; $0.202$/$0.362$, $p=0.001$), and PPA helps the most collapsed model ($0.362 \to 0.243$, $p=0.012$; Appendix~\ref{app:non-likert}).

\paragraph{Is sampling failing only where the model lacks knowledge?} A natural worry is that describe wins only on items the model happens to know well, and that where its distributional knowledge is poor, per-call sampling would be the safer choice. Two item-level analyses on the 100 OpinionQA items argue against this. First, each item's describe error and its persona-sampling error, both measured against Pew, are essentially uncorrelated ($\rho = 0.008$). Sampling does not fail because knowledge is missing, and knowing an item well does not rescue sampling. Second, on the 25 items the model describes worst, persona sampling is still the less accurate pipeline ($0.494$ vs.\ $0.339$; paired $p = 0.0018$). Poor knowledge therefore does not make sampling preferable, and describe remains more accurate than persona sampling on every population we test against ground truth, including the non-Likert items of Appendix~\ref{app:non-likert}.

\section{Prompt-Perturbed Argyle (PPA)}
\label{sec:ppa}

The describe pathway of \S\ref{sec:silicon} is the right choice when the application needs only a population estimate, but some applications require one simulated response per persona. Generative-agent simulations \citep{park2023generative, park2024generative} condition downstream reasoning on each agent's individual response, so a single-call describe pipeline cannot substitute. This section presents Prompt-Perturbed Argyle (PPA), a same-cost modification of standard Argyle for these applications.

\subsection{Method}
\label{sec:ppa-method}

Under standard Argyle, the same persona on the same item produces the same single answer in every call on most cells (\S\ref{sec:silicon-results}), so aggregating across calls cannot recover the within-stratum distribution. PPA perturbs the prompt's surface features across the 100 calls, holding the persona's content fixed.

PPA randomizes three surface features per call:

\begin{itemize}
\setlength{\itemsep}{2pt}
\item \textbf{Option ordering} (4 choices): ascending, descending, or one of two shuffled orderings of the Likert options.
\item \textbf{Question phrasing} (3 choices): direct, scale-anchored (``on a scale of 1 to 5\ldots''), or third-person.
\item \textbf{Prompt position} (3 choices): persona in system message, persona in user message, or persona in user message with brief recap.
\end{itemize}

Each call independently draws one combination uniformly at random from the $4 \times 3 \times 3 = 36$ possibilities. Standard Argyle uses a fixed canonical setting (ascending order, direct phrasing, persona in system message). Both pipelines make exactly 100 calls per item, one per persona, so the API cost is identical. Full protocol and surface-feature templates are in Appendix~\ref{app:ablation}.

Perturbing all three across calls dislodges the per-call anchor without changing what is asked. This extends \citet{dominguezolmedo2024questioning}: order sensitivity, applied across calls, becomes a tool for spreading each persona's responses. We test in \S\ref{sec:entropy-mechanism} whether the perturbations raise per-call entropy and whether that increase predicts the TV improvement.

\subsection{Results}
\label{sec:ppa-results}

\begin{table}[t]
\centering
\small
\resizebox{\columnwidth}{!}{%
\begin{tabular}{lcc}
\toprule
Pipeline & Mean TV & $\Delta$ vs.\ baseline \\
\midrule
Standard Argyle (baseline) & 0.46 & --- \\
PPA & \textbf{0.36} & $\mathbf{-0.10}$ \\
\bottomrule
\end{tabular}%
}
\caption{PPA vs.\ standard Argyle on 100 OpinionQA items, same persona set (100 personas per item). PPA reduces mean TV by 0.10 (21\% relative) at no added cost. Decoding-parameter sweeps (temperature, top-$p$) capture less than 20\% of PPA's effect; see Appendix~\ref{app:sweeps}.}
\label{tab:ppa}
\end{table}

PPA reduces mean TV by 0.10 (a 21\% relative reduction) at no added API cost (Wilcoxon paired $p = 1.7 \times 10^{-9}$). The effect is driven by the semantic perturbations rather than generic surface variation. On a pre-registered 13-item development subset of the 100 items, a control intervention that randomizes only whitespace, punctuation, and prompt prefix captures only 13\% of PPA's TV reduction (Appendix~\ref{app:ablation}).

\subsection{Mechanism: PPA works by raising per-call entropy}
\label{sec:entropy-mechanism}
\S\ref{sec:silicon-results} showed that standard Argyle returns the same single answer on nearly every call. PPA varies the prompt across calls, so the same persona can produce different answers, raising per-call entropy. We test whether PPA's TV improvement comes specifically from this entropy increase.

For each of 20 OpinionQA items (5 from each pre-registered cell), we measure per-(persona, item) entropy across 50 calls under standard Argyle ($H_{\text{std}}$) and under PPA ($H_{\text{PPA}}$). We use a denser protocol (5 personas per item, 50 calls per cell) because per-cell entropy needs many repeated calls per pair.

For each item $i$, we compute:
\begin{gather}
\Delta H_i = H_{\text{PPA},i} - H_{\text{std},i} \\
\Delta\text{TV}_i = \text{TV}_{\text{std},i} - \text{TV}_{\text{PPA},i}
\end{gather}
Positive $\Delta\text{TV}_i$ means PPA improves accuracy. To separate the entropy mechanism from item difficulty, we control for the per-item Pew entropy $H_{\text{Pew},i}$:
\begin{equation}
\Delta\text{TV}_i = \alpha + \beta_1 \Delta H_i + \beta_2 H_{\text{Pew},i} + \varepsilon_i
\label{eq:entropy-regression}
\end{equation}

\begin{table}[h]
\centering
\small
\resizebox{\columnwidth}{!}{%
\begin{tabular}{lcc}
\toprule
Predictor & Coefficient & 95\% CI \\
\midrule
$\Delta H$ (PPA entropy increase) & $0.27^{***}$ & $[0.14, 0.39]$ \\
$H_{\text{Pew}}$ (item ambiguity) & $-0.08$ (n.s.) & $[-0.29, 0.14]$ \\
\midrule
$R^2$ & 0.55 & \\
$N$ items & 20 & \\
\bottomrule
\end{tabular}%
}
\caption{OLS regression of per-item TV improvement on per-item per-call entropy increase, controlling for Pew per-item entropy. Each 1-nat increase in per-call entropy predicts a 0.27 reduction in TV ($p < 0.001$). Item difficulty is non-significant.}
\label{tab:entropy-regression}
\end{table}

Fitting Equation~\ref{eq:entropy-regression} across the 20 items yields a significant positive coefficient on $\Delta H$ (Table~\ref{tab:entropy-regression}): PPA's effect on TV tracks how much it raises per-call entropy, not how ambiguous the item is. Items where PPA increases per-call entropy by more than 0.7 nats show TV improvements of 0.15--0.25; items where PPA barely changes per-call entropy show no improvement. The raw correlation between $\Delta H$ and $\Delta\text{TV}$ is $r = 0.73$ ($p = 2 \times 10^{-4}$); bootstrap resampling confirms the $\Delta H$ coefficient is positive in 100\% of 1000 resamples (95\% CI $[0.18, 0.37]$).

Two follow-up tests rule out alternative explanations (Appendix~\ref{app:ppa-mechanism}). First, non-semantic perturbations of the prompt --- whitespace, punctuation, prefix variation --- raise per-call entropy by only $\Delta H = 0.003$ nats, two orders of magnitude smaller than PPA's $\Delta H = 0.417$, and yield no TV improvement at all ($\Delta\text{TV} = -0.002$ vs.\ $+0.112$ for PPA). What matters is the entropy increase, not prompt variation in general. Second, decoding temperature ($T = 1.5$) raises entropy less but yields a comparable $\Delta\text{TV}/\Delta H$ ratio (0.21 vs.\ 0.27; Table~\ref{tab:entropy-mechanism}): a similar improvement per nat of entropy emerges from two unrelated interventions.

\begin{table}[h]
\centering
\small
\resizebox{\columnwidth}{!}{%
\begin{tabular}{lccc}
\toprule
Intervention & $\Delta H$ (nats) & $\Delta\text{TV}$ & $\Delta\text{TV}/\Delta H$ \\
\midrule
PPA                  & 0.417 & $+0.112$ & 0.270 \\
\midrule
\multicolumn{4}{l}{\emph{Non-semantic controls}} \\
Whitespace + punct.\ + prefix & 0.003 & $-0.002$ & --- \\
\midrule
\multicolumn{4}{l}{\emph{Decoding-level analogue}} \\
Temperature $T=1.5$  & 0.069 & $+0.015$ & 0.212 \\
\bottomrule
\end{tabular}
}
\caption{Mechanism controls. Non-semantic perturbations raise per-call entropy by essentially zero and yield no TV improvement. Temperature raises entropy by less than PPA but produces a comparable TV improvement per nat ($\Delta\text{TV}/\Delta H \approx 0.21$ vs.\ $0.27$).}
\label{tab:entropy-mechanism}
\end{table}

\subsection{The induced variation is structured, not noise}
\label{sec:ppa-structure}

Against a collapsed baseline almost any added spread lowers TV, so the reduction alone does not show that PPA adds structure rather than noise. Three tests distinguish the two. First, noise would affect every persona alike and wash out subgroup differences; instead, by a per-item chi-square test, Democrat- and Republican-leaning personas differ on 33 of 50 items under PPA versus 27 under standard sampling, and single-answer items fall from 6 to 2 (Table~\ref{tab:ppa-subgroup}). Second, PPA's per-item Democrat--Republican gap tracks the Pew gap more closely ($r = 0.573 \to 0.678$; sign agreement $64\% \to 76\%$). The added variation moves the model toward real subgroup structure, not a common mixture (Appendix~\ref{app:ppa-subgroup}). Third, PPA's pooled estimate moves toward the distribution the same model reports under describe, not toward uniform: on the two open models, the TV between the pooled persona answers and the model's own described distribution falls from $0.507$ to $0.291$ (Qwen3-8B, $p = 2 \times 10^{-17}$) and from $0.556$ to $0.375$ (Qwen2.5-7B-Instruct, $p = 9 \times 10^{-11}$). In both cases the PPA estimate is closer to the described distribution than to uniform ($0.291$ vs.\ $0.360$; $0.375$ vs.\ $0.448$; Table~\ref{tab:ppa-convergence}). 

\paragraph{What remains open.} These results establish that PPA's variation is structured and that it improves subgroup fidelity, but not why re-ordering options or rephrasing a question changes the top answer of an otherwise deterministic model. The distributional knowledge is present in the base (\S\ref{sec:failure-alignment}), instruction-targeted post-training collapses the per-call read-out, and surface perturbation partly releases it; why the read-out is order-sensitive is left to future work.

\subsection{Open-ended generation}
\label{sec:ppa-openended}

Re-ordering does not exist outside multiple choice, so open-ended tasks isolate PPA's paraphrase component. On five open-ended survey tasks across four models (Llama-3.1-8B-Instruct, Qwen2.5-7B-Instruct, Qwen3-8B, OLMo-2-7B-Instruct; 200 calls per condition, $T=1.0$), paraphrase-only PPA raises per-call entropy by $0.98$ nats on average, while non-semantic surface variation of the same prompt raises it by $0.01$. The gain is largest where the fixed prompt is most collapsed ($\rho=-0.75$). The failure and its partial repair are not Likert-specific (Appendix~\ref{app:openended}).

\section{Discussion}
\label{sec:discussion}

We have documented a per-call sampling failure in instruction-tuned language models, traced it to instruction-targeted post-training through a base-vs-instruct comparison, and shown that the model's distributional knowledge remains accessible through a describe-style interface. Applying the framework to silicon sampling, we have provided two mechanism-grounded mitigations evaluated on 100 OpinionQA items.

\paragraph{Relation to mode-collapse and decoding.} Mode-collapse research documents reduced diversity in RLHF-aligned models \citep{padmakumar2024diversity} with causes framed at training time \citep{ouyang2022training, bai2022constitutional}. We refine the picture: the collapse is categorical rather than statistical (\S\ref{sec:failure-categorical}) and is amplified by instruction-targeted post-training (\S\ref{sec:failure-alignment}), consistent with the formal-vs-functional dissociation of \citet{mahowald2024dissociating}. A deterministic network is not a randomness source, and silicon sampling never asks it to be: decoding supplies the randomness, the model supplies a calibrated conditional. Instruction tuning warps that conditional, driving the top-two logit gap above 14 nats (Appendix~\ref{app:logit}), which would need a decoding temperature near 17 against an API maximum of 2. The failure is calibration, not randomness, and temperature resamples a distribution that is already collapsed.

\paragraph{Implications for other LLM applications.} The per-call assumption is inherited by several other LLM applications. Generative-agent simulations \citep{park2023generative, park2024generative} treat each agent's per-turn response as a draw from a distribution; if each agent instead collapses to a fixed answer, the simulation may understate emergent tail behaviors. Synthetic data generation pipelines \citep{wang2022self, honovich2022unnatural} and automated dataset construction \citep{liu2022wanli} treat per-call outputs as samples from a target, and LLM-as-judge protocols assume that ratings across calls are independent draws. For instruction-tuned models, distributional fidelity should be verified rather than assumed.

\paragraph{Practical guidance.} When the application needs only a population estimate, a describe pipeline is preferable: a single call returning a JSON distribution is roughly $2.1\times$ more accurate than persona aggregation on OpinionQA, and the advantage holds against demographically-matched baselines. When per-respondent outputs are required and a describe pipeline cannot substitute, PPA reduces TV by 21\% at no added API cost through randomization of option ordering and question phrasing.

\paragraph{Toward a training-time fix.} Our mitigations act at inference time and leave the underlying primitive untouched. Because the damage tracks instruction-targeted post-training (\S\ref{sec:failure-alignment}) and accrues at every stage, a durable fix likely belongs in the training objective rather than in any single stage. Three directions follow from our findings. First, a loss that matches the stated output distribution when an instruction specifies one, rather than rewarding a single preferred answer. Second, rewards scored over groups of samples, since single-answer rewards make repeating one best answer optimal and thereby select for collapse. Third, evaluation on repeated-call distributions, since single-call evaluation cannot detect the failure at all. We do not run these experiments here; the fix belongs in the objectives that shape the per-call read-out.

\section{Conclusion}

Instruction-tuned language models do not sample from distributions; they collapse to a single output that survives temperature, top-$p$, and prompt-level corrections. Yet the same models can describe those distributions accurately in a single call, and our base-vs-instruct comparison shows that the per-call sampling failure is amplified by instruction-targeted post-training rather than inherent to the architecture. For applications that need population estimates, asking the model to describe the response distribution directly more than halves the error against human survey data; for applications that need per-respondent outputs, Prompt-Perturbed Argyle reduces the same error by 21\% at no added cost. The broader implication is that any LLM pipeline that treats per-call outputs as samples from a target distribution should verify that assumption rather than inherit it.

\section*{Acknowledgements}
This work was supported by the National Research Foundation of Korea (NRF) grant (Grant No. RS-2025-00563196) and the Institute of Information \& Communications Technology Planning \& Evaluation (IITP) grants (Grant Nos. RS-2019-II191126 and RS-2024-0045974) funded by the Korean government (MSIT). This work was also supported by the AI Science Hub Program of the Ministry of Science and ICT (Grant No. N10260126).

\newpage
\section*{Limitations}

\paragraph{Sampling-as-output tasks only.} Our findings concern whether the model's output distribution across many calls matches a target. Per-token correctness tasks (translation, summarization, factual recall) are out of scope.

\paragraph{Response formats tested.} We establish the collapse, the describe advantage, and PPA's repair on 5-point Likert, 2--4 option, and five open-ended tasks. Whether the same pattern holds on structured generation, long-form answers, or non-survey tasks is left to future work.

\paragraph{Ceiling of both mitigations.} Both fixes are bounded by the model's prior. Describe's TV-to-Pew is a same-session ceiling that may degrade on populations outside the training distribution, and PPA can only redistribute mass within the prior, not add options the prior excludes. Our recommendation rests on comparisons against ground truth, where describe dominates sampling even on the items it describes worst (\S\ref{sec:silicon-results}); for a new population with no ground truth, describe accuracy cannot be verified in advance.

\paragraph{Entropy-correlation scope.} The entropy-correlation regression establishes PPA's per-call mechanism on 20 items; describe is a single-call method for which per-call entropy is undefined, so the same regression cannot demonstrate its mechanism.

\paragraph{Scale and lineage.} The base-vs-instruct comparison uses 7--8B models, where clean base--instruct pairs are available; the cross-family panel reaches frontier-scale instruct models but cannot supply matched bases, and it conflates size with generation. The direction is established at 7--8B and replicates on the instruct side at frontier scale; a matched comparison at 70B+ would strengthen the causal claim.

\paragraph{Component-level attribution.} Our staging and continued-pretraining results (\S\ref{sec:failure-alignment}) localize the effect to instruction-targeted post-training rather than to additional training in general, but they do not isolate which component is responsible. The OLMo-2 stages are cumulative rather than factorial, so they show where damage accrues, not the independent contribution of SFT, preference optimization, or RL. Disentangling the per-component effect is left to future work.

\section*{Ethical Considerations}

\paragraph{Data.} We use publicly available Pew ATP data \citep{pew_atp} via OpinionQA \citep{santurkar2023whose}; no human-subjects research was conducted.

\paragraph{Artifacts.} All models and datasets are used under their respective licenses; our released prompts, code, and benchmark results are for research use.

\paragraph{Risks.} LLM-simulated populations should not substitute for real human survey data in policy-relevant settings; the describe pathway and PPA mitigate but do not eliminate the failure, and application-specific validation against human ground truth remains necessary.

\paragraph{AI assistants.} We used AI coding assistants and writing-editing tools during preparation; all experimental design, analysis, and conclusions are the authors' own.

\bibliography{custom}

\appendix

\FloatBarrier
\section{Experimental setup}
\label{app:setup}

\paragraph{Main model.} \texttt{gpt-4o} accessed via the OpenAI Chat Completions API. The cross-family replication (Appendix~\ref{app:cross-family}) uses additional API endpoints (\texttt{gpt-5.4}, \texttt{claude-sonnet-4.5}, \texttt{gemini-3-flash-preview}, \texttt{deepseek-chat}) and the open-weight \texttt{llama-3.3-70b-instruct} accessed via a hosted inference API.

\paragraph{Sampling.} Behavioral sampling uses temperature 1.0, top-$p$ 1.0, max-tokens 50, with $N = 200$ calls per condition. Logit reads use temperature 0 and the API's \texttt{top\_logprobs} parameter set to 20.

\paragraph{Personas.} We use 100 personas sampled with replacement from the joint distribution of six demographic attributes (race, sex, age, education, income, party) in Pew ATP Wave 92, rendered with the fixed one-sentence template of \citet{argyle2023out}. The same persona set is used for all standard-Argyle OpinionQA experiments (baseline, PPA, decoding-parameter sweeps, ablations). Matched-Argyle uses per-item demographic-matched personas as described in Appendix~\ref{app:matched-argyle}.

\paragraph{Items.} The 100 OpinionQA items are drawn deterministically by item-key sort, restricted to 5-point Likert with usable ground truth. The first 50 items come from the primary wave list (Pew ATP Waves 26, 27, 34, 42, 43, 50, 54). The second 50 items extend the wave list (in pre-committed order: W92, W36, W41, W49, W82, W45, W32). The full item list, the FACTS/VALUES annotation, and the entropy-bin labels are committed to the repository.

\paragraph{Parsing.} For Likert tasks, we accept any prefix-numeric response in \{1, 2, 3, 4, 5\} and a small set of synonym strings (``strongly agree''$\to$5, etc.); unparseable responses are excluded from analysis. For RNG tasks, we accept only the bare integer/letter/float and reject any chain-of-thought, free text, or out-of-range value.

\paragraph{Statistics.} For a discrete sample space $\mathcal{X}$ and two distributions $P, Q$ over $\mathcal{X}$, the total-variation (TV) distance is
\begin{equation*}
\mathrm{TV}(P, Q) = \frac{1}{2} \sum_{x \in \mathcal{X}} \lvert P(x) - Q(x) \rvert .
\end{equation*}
TV is bounded in $[0, 1]$, with $0$ indicating perfect match and $1$ indicating disjoint support. We use TV because it is symmetric (unlike KL, which diverges when the target assigns zero probability to a covered option) and bounds the maximum difference in event probabilities between $P$ and $Q$.

In all experiments, we report TV between the empirical distribution over $N$ model calls and the target distribution specified by the prompt (for the synthetic targets of \S\ref{sec:failure}) or by the Pew ground truth (for OpinionQA in \S\ref{sec:silicon}).

\paragraph{Prompt templates.} Verbatim prompt text for every condition is in the attached repository; see Appendix~\ref{app:correction-prompts} for the seven DOES corrections and two KNOWS interventions used in \S\ref{sec:failure-knowsdoes}.

\FloatBarrier
\section{Base-vs-instruct protocol}
\label{app:base-instruct}

\paragraph{Models.} Tables~\ref{tab:base-instruct}, \ref{tab:staging}, \ref{tab:base-instruct-v8} and~\ref{tab:knockout} come from one local protocol over \texttt{meta-llama/Llama-3.1-8B} / \texttt{-Instruct}, \texttt{mistralai/Mistral-7B-v0.3} / \texttt{Mistral-7B-Instruct-v0.3}, \texttt{Qwen/Qwen2.5-7B} / \texttt{-Instruct}, \texttt{Qwen/Qwen2.5-Coder-7B}, and \texttt{allenai/OLMo-2-1124-7B}, \texttt{-SFT}, \texttt{-DPO}, \texttt{-Instruct} (RLVR), on a single A100 GPU, seed 13 (seed 47 replicates every ordering).

\paragraph{Prompting.} Base models do not reliably follow zero-shot instructions, so we use a 3-shot in-context format for V0 on the base model: three (Q, A) example pairs followed by the test question, with one canonical Q/A example for V8. The instruct model uses the standard chat template (zero-shot). This separates 'cannot follow the instruction' from 'cannot sample.'

\paragraph{Sampling.} Temperature 1.0, top-$p$ 1.0, 200 V0 calls per task, 10 V8 calls per task (each returning a list of 20). Per family: 2 models $\times$ 5 tasks $\times$ 200 V0 calls $= 2{,}000$ single-answer generations. V8 (Table~\ref{tab:base-instruct-v8}): 2 models $\times$ 5 tasks $\times$ 10 calls $= 100$ list generations (2{,}000 list entries).

\paragraph{V8 results.} Table~\ref{tab:base-instruct-v8} reports V8 (list-of-$N=20$) TV-to-target for the base and instruct models on the same five tasks. The base model recovers the target to within 0.10 on three of five tasks and stays below 0.20 on all five; the instruct model is within 0.10 on all five. This rules out a ``base model cannot follow the V8 instruction'' confound and shows that the KNOWS pathway is already present pre-alignment.

\begin{table}[ht]
\centering
\small
\resizebox{\columnwidth}{!}{%
\begin{tabular}{lcc}
\toprule
Task & Base V8 TV & Instruct V8 TV \\
\midrule
uniform 1--5   & 0.05 & 0.02 \\
fair coin      & 0.01 & 0.05 \\
skewed binary  & 0.07 & 0.06 \\
bimodal 5-way  & 0.17 & 0.10 \\
skewed 5-way   & 0.11 & 0.06 \\
\bottomrule
\end{tabular}
}
\caption{V8 (list-of-$N{=}20$) TV-to-target on \texttt{Llama-3.1-8B} base vs.\ instruct, local replication. Both models recover the target to within 0.10 on three or more tasks. The KNOWS pathway is present pre-alignment.}
\label{tab:base-instruct-v8}
\end{table}

\paragraph{Why 8B vs.\ a larger model.} The 8B model fits comfortably on a single A100. Comparable base-vs-instruct pairs are not as cleanly available at 70B (different fine-tuning lineages). The 7--8B result establishes the qualitative direction; scaling within Llama-3 (e.g., 70B base vs.\ instruct, if released) would refine it.

\paragraph{Update scaling (task arithmetic).} For two lineages with a released base and its instruction-tuned successor, we form $\tau = \theta_{\text{tuned}} - \theta_{\text{base}}$ and evaluate $\theta_{\text{base}} + \alpha\tau$ at $\alpha \in \{0, 0.25, 0.5, 0.75, 1, 1.25\}$ \citep{ilharco2023editing}: $\alpha = 0$ is the base checkpoint, $\alpha = 1$ the released tuned checkpoint, and $\alpha = 1.25$ extrapolates past it. Every $\alpha$ is run with the same base-style 3-shot prompt (V0, the five-target panel, 200 calls per task at $T = 1.0$, seed 13), so only the weights vary. Because the tuned checkpoint is therefore prompted in the base format rather than through its chat template, its $\alpha = 1$ mean differs from the chat-template numbers of Table~\ref{tab:staging} (0.317 vs.\ 0.321 for OLMo-2 SFT; 0.459 vs.\ 0.540 for Qwen2.5-Instruct); the sweep is a separate run, so its base means also differ slightly (Qwen2.5: 0.198 vs.\ 0.203). Alongside sampling TV we read, at $T = 0$, the probability the model assigns to its modal answer over the task's support, averaged over the five tasks.

\begin{table}[ht]
\centering
\small
\resizebox{\columnwidth}{!}{%
\begin{tabular}{lcccccc}
\toprule
$\alpha$ & 0 & 0.25 & 0.5 & 0.75 & 1 & 1.25 \\
\midrule
\multicolumn{7}{l}{\emph{OLMo-2 base $\to$ SFT}} \\
Mean TV-to-target        & 0.236 & 0.234 & 0.254 & 0.274 & 0.317 & 0.410 \\
$T{=}0$ modal probability & 0.395 & 0.404 & 0.431 & 0.456 & 0.477 & 0.550 \\
\midrule
\multicolumn{7}{l}{\emph{Qwen2.5 base $\to$ Instruct}} \\
Mean TV-to-target        & 0.198 & 0.248 & 0.291 & 0.379 & 0.459 & 0.519 \\
$T{=}0$ modal probability & 0.588 & 0.609 & 0.635 & 0.702 & 0.849 & 0.947 \\
\bottomrule
\end{tabular}%
}
\caption{Scaling the instruction-tuning update. Mean V0 TV-to-target over the five-target panel and mean $T{=}0$ probability of the modal answer at $\theta_{\text{base}} + \alpha\tau$, prompt held fixed (200 calls per task, seed 13). Both rise with $\alpha$ on both lineages; $\alpha = 0$ reproduces the base checkpoint.}
\label{tab:knockout}
\end{table}

\FloatBarrier
\section{Demographically-matched Argyle protocol}
\label{app:matched-argyle}
\paragraph{Describe vs.\ Verbalized Sampling.} Verbalized Sampling \citep{zhang2025verbalized} prompts the model to verbalize the response distribution as JSON probabilities in a single call. Our describe pathway uses the same query structure. On the 100 OpinionQA items, the two pipelines produce statistically indistinguishable TV-to-Pew (Describe: 0.22; VS: 0.23; Wilcoxon paired $p = 0.14$, $n = 100$). We use the describe pathway in the main results because it integrates with the broader KNOWS/DOES framework, but the underlying mechanism — single-call distributional query — is the same.

\paragraph{Goal.} Control for persona-prior mismatch in the describe-vs-Argyle comparison. The standard Argyle baseline uses 100 generic personas drawn from the Wave-92 joint demographic distribution reused across all items. A skeptic could argue describe wins because these personas don't match the actual Pew respondent demographic distribution for each item. Matched Argyle eliminates this confound by drawing personas from each item's actual respondent demographic distribution.

\paragraph{Per-item respondent pools.} For each of the 100 OpinionQA items, we identify the Pew ATP wave it comes from, load the wave's respondent-level data from the OpinionQA release \citep{santurkar2023whose}, and keep the respondents who answered the item (non-null response).

\paragraph{Persona sampling.} For each item, we sample 100 actual respondent rows (with replacement, using the wave's survey weights) from the Pew respondents who answered that item and render each with the six-attribute template of Appendix~\ref{app:setup}, using the same persona-formatting code as standard Argyle, so the only difference from standard Argyle is which personas are used. Matching therefore operates on the full joint distribution of respondents, while the persona text carries the same six attributes as standard Argyle.

\paragraph{API calls and aggregation.} For each (matched persona, item) pair: a single Argyle call at temperature 1.0 using \texttt{gpt-4o}. Total: 100 items $\times$ 100 matched personas $= 10{,}000$ calls. Aggregate to per-item LLM response distribution; compute TV against the Pew per-item distribution.

\paragraph{Result.} Matched Argyle mean TV $= 0.484$, vs.\ standard Argyle TV $= 0.459$. The two are not significantly different (paired Wilcoxon $p = 0.092$, $n = 100$ items); if anything matched Argyle is slightly worse, suggesting that matching personas to demographics slightly amplifies the model's tendency to stereotype the demographic group. Describe still beats matched Argyle by $\Delta\text{TV} = 0.269$ ($p = 4.5 \times 10^{-16}$).

\paragraph{Open-model replication.} The 100-item comparison holds on three open 7--8B instruct models run locally (Table~\ref{tab:openmodel-main}): describe beats standard Argyle on all three, and PPA closes 63--72\% of the Argyle-to-describe gap on the Qwen models and edges past describe on Llama (0.276 vs.\ 0.293).

\begin{table}[ht]
\centering
\small
\begin{tabular}{lccc}
\toprule
Model & Argyle & PPA & Describe \\
\midrule
Llama-3.1-8B-Instruct & 0.436 & 0.276 & 0.293 \\
Qwen2.5-7B-Instruct   & 0.525 & 0.345 & 0.241 \\
Qwen3-8B              & 0.465 & 0.304 & 0.240 \\
\bottomrule
\end{tabular}
\caption{Standard Argyle, PPA, and describe on the 100 OpinionQA items, three open models, 100 persona calls per item (describe: one call per item; mean over the 96, 82, and 100 items for which the model returned a distribution). Describe vs.\ Argyle, paired Wilcoxon: $p = 1.3 \times 10^{-8}$, $7.6 \times 10^{-14}$, $3.5 \times 10^{-13}$.}
\label{tab:openmodel-main}
\end{table}

\FloatBarrier
\section{Per-cell entropy on OpinionQA}
\label{app:per-cell-entropy}

For 20 OpinionQA items (5 from each pre-registered cell) and 5 personas per item, we run standard Argyle 50 times per (persona, item) cell. Per-cell entropy statistics across the 100 cells:

\begin{table}[ht]
\centering
\small
\resizebox{\columnwidth}{!}{%
\begin{tabular}{lc}
\toprule
Metric & Value \\
\midrule
Cells with entropy $= 0$ (perfectly deterministic) & 57/100 \\
Cells with top response in $> 90\%$ of calls & 80/100 \\
Cells with top response in $> 95\%$ of calls & 75/100 \\
Mean per-cell entropy & 0.145 nats \\
Mean per-cell entropy as $\%$ of max & 9.0\% \\
Median per-cell entropy & 0.000 \\
\bottomrule
\end{tabular}%
}
\caption{Per-cell entropy across 100 (persona, item) cells under standard Argyle, 50 calls per cell. The \S\ref{sec:failure}--\S\ref{sec:silicon} categorical collapse replicates on real OpinionQA items.}
\label{tab:per-cell-entropy}
\end{table}

On real opinion items under standard persona-conditioned Argyle, more than half of the (persona, item) cells produce a perfectly deterministic response across 50 repeated calls.

\FloatBarrier
\section{Seven-target panel}
\label{app:seven-targets}

Per-target results at $N = 200$ calls each, $T = 1.0$.

\begin{table}[ht]
\centering
\small
\resizebox{\columnwidth}{!}{%
\begin{tabular}{lcccc}
\toprule
Target & target distribution & P(mode) & TV & $N$ \\
\midrule
point (always 5)            & $[0,0,0,0,1]$               & 1.000 & 0.000 & 200 \\
mixture $5\text{:}70\%/4\text{:}30\%$ & $[0,0,0,0.3,0.7]$  & 1.000 & 0.300 & 200 \\
skewed binary $0.7/0.3$     & $[0.7, 0.3]$                & 1.000 & 0.300 & 200 \\
bimodal $1$ \& $5$, never $2/3/4$ & $[0.5, 0, 0, 0, 0.5]$ & 1.000 & 0.500 & 200 \\
three-way $0.5/0.3/0.2$     & $[0.5, 0.3, 0.2]$           & 0.980 & 0.480 & 200 \\
skewed 5-way                & $[0.40, 0.30, 0.15, 0.10, 0.05]$ & 1.000 & 0.600 & 200 \\
uniform 1--5                & $[0.2, 0.2, 0.2, 0.2, 0.2]$ & 0.943 & 0.743 & 200 \\
\bottomrule
\end{tabular}%
}
\caption{Seven-target panel results. The model produces a single answer in over 94\% of calls on every non-point target.}
\label{tab:seven-targets}
\end{table}

Mean $P(\text{mode})$ across the six non-point targets is 0.987.

\paragraph{Swap test.} Repeating the skewed binary target with options A and B swapped:

\begin{table}[ht]
\centering
\small
\resizebox{\columnwidth}{!}{%
\begin{tabular}{lccc}
\toprule
Condition & A & B & mode \\
\midrule
$[A=0.7, B=0.3]$        & 200/200 & 0/200   & A \\
$[A=0.3, B=0.7]$ (swap) & 0/200   & 200/200 & B \\
no-instruction baseline & 200/200 & 0/200   & A \\
\bottomrule
\end{tabular}%
}
\caption{Swap test on skewed binary. The model relocates its single output but never produces a distribution.}
\label{tab:swap-test}
\end{table}

The swap relocates the modal output (A $\to$ B) but never produces the $[0.3, 0.7]$ shape; the model treats the instruction as identifying a single answer.

\FloatBarrier
\section{Logit-level results}
\label{app:logit}

At temperature 0 with \texttt{top\_logprobs}=20, restricted to relevant outputs and renormalized. Top-2 logit gap (in nats) on the four non-point targets, with and without the distribution instruction.

\begin{table}[ht]
\centering
\small
\resizebox{\columnwidth}{!}{%
\begin{tabular}{lrrr}
\toprule
Target & no instruction & with instruction & widening \\
\midrule
skewed binary $[0.7, 0.3]$ & 6.50 & 14.75 & $+8.25$ \\
three-way $[0.5, 0.3, 0.2]$ & --- & 3.75 & --- \\
bimodal $[0.5, 0, 0, 0, 0.5]$ & --- & 9.21 & --- \\
skewed 5-way $[0.40, 0.30, 0.15, 0.10, 0.05]$ & 0.00 & $\geq 16$ & $\geq +16$ \\
\bottomrule
\end{tabular}%
}
\caption{Top-2 logit gap (nats) on non-point targets at $T=0$. Adding the distribution instruction widens the gap rather than narrowing it.}
\label{tab:logit-gap}
\end{table}

Within-support $P(\text{mode})$ at $T = 0$ across all four with-instruction conditions: $\geq 0.976$. The skewed 5-way case requires sampling temperature $\geq 16 / \log(0.40/0.30) \approx 56$ to recover the target, far above the API maximum of 2.

\FloatBarrier
\section{RNG primitive across formats}
\label{app:rng}

Five conditions probing the model's ability to produce randomness directly, no persona or task framing. $N = 200$ calls per condition at $T = 1.0$.

\begin{table}[ht]
\centering
\small
\resizebox{\columnwidth}{!}{%
\begin{tabular}{llcccc}
\toprule
Condition & Support & Modal value & $P(\text{mode})$ & Coverage & $p$ vs.\ uniform \\
\midrule
R1 int 1--100  & $\{1,\ldots,100\}$ & 42 & 0.775 & 15/100 & $\chi^2 = 11{,}971$, $< 10^{-300}$ \\
R2 int 1--10   & $\{1,\ldots,10\}$  & 7  & 0.970 & 4/10   & $\chi^2 = 1{,}683$, $< 10^{-300}$ \\
R3 float [0,1] & $[0, 1]$           & \multicolumn{2}{c}{mean 0.552, std 0.108, range [0.27, 0.76]} & n/a & KS $D = 0.370$, $5 \times 10^{-25}$ \\
R4 coin H/T    & $\{H, T\}$         & H  & 0.854 & 2/2    & binomial, $2 \times 10^{-25}$ \\
R5 letter A--Z & $\{A,\ldots,Z\}$   & G  & 0.465 & 16/26  & $\chi^2 = 1{,}183$, $7 \times 10^{-234}$ \\
\bottomrule
\end{tabular}%
}
\caption{RNG primitive across five output formats. Every format rejects uniformity at $p < 0.001$; the model concentrates on a format-specific favorite.}
\label{tab:rng-primitive-full}
\end{table}

\paragraph{Exclusion conditions.} For each of the three discrete cases, we re-ran the probe with the modal value excluded by name. The model honored the exclusion in every case (no violations) but relocated mass onto a different favored value.

\begin{table}[ht]
\centering
\small
\resizebox{\columnwidth}{!}{%
\begin{tabular}{lccccc}
\toprule
Condition & Violations & New mode & $P(\text{new mode})$ & Coverage (remaining) & TV vs.\ uniform \\
\midrule
C1 exclude 42 & 0/200 & 37 & 0.520 & 20/99 & 0.854 \\
C2 exclude 7  & 0/200 & 3  & 0.710 & 7/9   & 0.663 \\
C3 exclude G  & 0/200 & K  & 0.155 & 22/25 & 0.485 \\
\bottomrule
\end{tabular}%
}
\caption{Exclusion conditions: the model honors the exclusion but relocates mass onto a different favorite. The collapse is structural, not lexical.}
\label{tab:rng-exclude}
\end{table}

\FloatBarrier
\section{Correction-strategy prompts (V0--V9)}
\label{app:correction-prompts}

\paragraph{DOES interventions (V0--V7).}
\begin{itemize}
\item \textbf{V0 plain sampling.} The bare task prompt at $T = 1.0$, no correction.
\item \textbf{V1 verbatim feedback.} A system prompt that reports the model's measured behavior on the task prompt (``Across 200 runs, you produced: digit 3: 94.5\%\ldots\ This is far from the target'') and asks it to produce the target distribution.
\item \textbf{V2 apologetic framing.} A system prompt in which the \emph{user} apologizes for unclear earlier instructions and restates the target.
\item \textbf{V3 third-person framing.} The measured behavior and the target are stated about ``the assistant'' in the third person.
\item \textbf{V4 explicit bias-naming.} The prompt names mode collapse directly and instructs the model to counteract it.
\item \textbf{V5 in-context demonstration.} 20 shuffled exemplars matching the target distribution are provided before the task prompt.
\item \textbf{V6 step-by-step reasoning.} A four-step scaffold (acknowledge the target, identify the support, mentally simulate a draw, report it) applied silently; the user message ends ``Apply the steps above silently and respond with only the result.''
\item \textbf{V7 algorithmic chain-of-thought.} An inverse-CDF scaffold: state the cumulative distribution, draw $u \in [0,1]$, output the option whose cumulative range contains $u$.
\end{itemize}

\paragraph{KNOWS interventions (V8--V9).}
\begin{itemize}
\item \textbf{V8 list-of-$N$.} A single call returns $N$ exemplars whose empirical distribution should match the target.
\item \textbf{V9 parametric specification.} A single call returns the target distribution as JSON probabilities; an external library performs the sampling.
\end{itemize}

\FloatBarrier
\section{KNOWS/DOES per-task results}
\label{app:knows-does}

TV-to-target for each intervention $\times$ task. V0--V7: $N = 200$ calls per cell at $T = 1.0$; V8: 10 calls each returning 20 samples; V9: one call at $T = 0$ returning a distribution, from which 200 draws are taken externally (NumPy, seed 42). All on \texttt{gpt-4o}. Full intervention text is in Appendix~\ref{app:correction-prompts}.

\begin{table}[ht]
\centering
\small
\resizebox{\columnwidth}{!}{%
\begin{tabular}{lccccc}
\toprule
Variant & uniform 1--5 & fair coin & skewed binary & bimodal 5-way & skewed 5-way \\
\midrule
V0 plain sampling   & 0.743 & 0.475 & 0.300 & 0.600 & 0.500 \\
V1 verbatim         & 0.655 & 0.305 & 0.300 & 0.600 & 0.500 \\
V2 apologetic       & 0.685 & 0.295 & 0.300 & 0.600 & 0.500 \\
V3 third-person     & 0.590 & 0.500 & 0.300 & 0.600 & 0.500 \\
V4 explicit-bias    & 0.580 & 0.500 & \textbf{0.125} & 0.780 & 0.600 \\
V5 20-shot ICL      & 0.650 & 0.485 & 0.300 & 0.600 & 0.460 \\
V6 reasoning        & 0.700 & 0.110 & 0.300 & 0.520 & 0.500 \\
V7 CoT-explicit-prob& 0.700 & 0.435 & 0.300 & 0.460 & 0.490 \\
\midrule
V8 list-of-$N{=}20$ & \textbf{0.000} & \textbf{0.045} & \textbf{0.050} & \textbf{0.050} & \textbf{0.000} \\
V9 JSON probabilities     & \textbf{0.045} & \textbf{0.020} & \textbf{0.010} & \textbf{0.025} & \textbf{0.035} \\
\bottomrule
\end{tabular}%
}
\caption{TV-to-target for every intervention $\times$ task. Bold cells indicate TV $\leq 0.05$ (the KNOWS threshold). Every KNOWS cell is at or below the threshold; no DOES cell is.}
\label{tab:knows-does-full}
\end{table}

The separation is clean at the per-task level, not just on the across-task mean reported in \S\ref{sec:failure-knowsdoes}. Across $8 \times 5 = 40$ DOES cells, the minimum TV is $0.110$ (V6 on the fair coin) and only 2/40 fall below 0.20 (V6 on the fair coin at 0.110 and V4 on the skewed binary at 0.125), and both remain more than twice the largest KNOWS cell. Across the $2 \times 5 = 10$ KNOWS cells, the maximum TV is $0.050$, and 5/10 are at or below $0.025$. The gap is not a property of any single task: every task contributes the same qualitative ordering, with both KNOWS variants strictly dominating all eight DOES variants on that task.

\FloatBarrier
\section{CoT-RNG bridge analysis}
\label{app:cot-bridge}

We test whether V7's failure inherits from the degenerate primitive of \S\ref{sec:failure-rng} by parsing the model's stated $u$ values from the V7 traces. 1000 responses across 5 tasks (200 each).

\begin{table}[ht]
\centering
\small
\resizebox{\columnwidth}{!}{%
\begin{tabular}{lrrrr}
\toprule
Task & $u$ mean & $u$ std & $u$ range & KS vs $U[0,1]$ \\
\midrule
uniform 1--5  & 0.465 & 0.166 & [0.12, 0.87] & $p = 2.3 \times 10^{-13}$ \\
fair coin      & 0.451 & 0.185 & [0.12, 0.74] & $p = 1.4 \times 10^{-17}$ \\
skewed binary  & 0.460 & 0.059 & [0.23, 0.65] & $p = 6.4 \times 10^{-31}$ \\
bimodal 5-way  & 0.349 & 0.144 & [0.12, 0.82] & $p = 9.3 \times 10^{-47}$ \\
skewed 5-way   & 0.370 & 0.107 & [0.12, 0.65] & $p = 2.7 \times 10^{-33}$ \\
\bottomrule
\end{tabular}%
}
\caption{V7 CoT-internal $u$ distribution per task. $u$ is non-uniform on every task.}
\label{tab:bridge-u}
\end{table}

\paragraph{Bridge $u \to$ answer agreement.} For each task we apply the inverse-CDF map to the model's stated $u$ and ask whether the resulting answer matches the answer the model actually produced. Table~\ref{tab:bridge-agreement} reports the per-task agreement rate.

\begin{table}[ht]
\centering
\small
\begin{tabular}{lr}
\toprule
Task & bridge agreement \\
\midrule
uniform 1--5  & 37.5\% \\
fair coin      & 99.5\% \\
skewed binary  & 100.0\% \\
bimodal 5-way  & 97.5\% \\
skewed 5-way   & 20.0\% \\
\bottomrule
\end{tabular}
\caption{Bridge $u \to$ answer agreement per task. Three tasks execute the inverse-CDF mapping on a restricted $u$; two write $u$ and ignore it.}
\label{tab:bridge-agreement}
\end{table}

\FloatBarrier
\section{V8 N-sweep}
\label{app:v8-nsweep}

V8 wraps the same query in a single call that requests a list of $N$ samples. We sweep $N \in \{1, 2, 5, 10, 20\}$ on four categorical tasks (40 calls per cell at $T = 1.0$, \texttt{gpt-4o}) to test whether the list-construction effect depends on list length.

\begin{table}[ht]
\centering
\small
\resizebox{\columnwidth}{!}{%
\begin{tabular}{lrrrrr}
\toprule
Task & $N{=}1$ & $N{=}2$ & $N{=}5$ & $N{=}10$ & $N{=}20$ \\
\midrule
uniform 1--5  & 0.60 & 0.24 & 0.00 & 0.01 & 0.01 \\
fair coin      & 0.45 & 0.00 & 0.06 & 0.00 & 0.00 \\
skewed binary  & 0.30 & 0.06 & 0.09 & 0.05 & 0.08 \\
skewed 5-way   & 0.50 & 0.30 & 0.20 & 0.01 & 0.01 \\
\bottomrule
\end{tabular}%
}
\caption{V8 TV-to-target across list lengths. V8 reverts to single-answer failure at $N = 1$; by $N = 5$ every task is at or below 0.20. Separate run from Table~\ref{tab:knows-does-full} (40 lists per cell).} 
\label{tab:v8-nsweep}
\end{table}

At $N = 1$, V8 collapses to ordinary per-call sampling and the failure returns in full: every task shows TV $\geq 0.30$, matching the V0--V7 range. Going from $N = 1$ to $N = 2$ already cuts TV by roughly half on three of four tasks. By $N = 5$ every task is at or below 0.20, and by $N = 20$ four of four are at or near zero. The transition is sharp, not gradual. More independent draws would predict $1/\sqrt{N}$ shrinkage in noise with no change in mean; instead V8 switches from per-call sampling to single-call counting once it has enough slots to lay out a recognizable shape (\S\ref{sec:failure-knowsdoes}).

\FloatBarrier
\section{PPA ablation}
\label{app:ablation}

The PPA ablation uses a 13-item pre-registered development subset of the 100 OpinionQA items. The subset was selected before any PPA results were computed (item keys committed in the released code). We use this smaller set for ablations because each variant requires a separate run with 100 personas per item; the development subset keeps the API budget tractable while preserving the same persona protocol as the main experiment.

\begin{table}[ht]
\centering
\small
\resizebox{\columnwidth}{!}{%
\begin{tabular}{lccc}
\toprule
Condition & Mean TV & $\Delta$ & \% of PPA \\
\midrule
\multicolumn{4}{l}{\emph{(a) Single-knob ablation}} \\
A0 baseline                        & 0.481 & ---      & --- \\
A1 order-only                      & 0.416 & $-0.064$ & 60\% \\
A2 phrasing-only                   & 0.422 & $-0.059$ & 56\% \\
\midrule
\multicolumn{4}{l}{\emph{(b) Cumulative ablation}} \\
A0 baseline                        & 0.481 & ---      & --- \\
$+\,$order                         & 0.416 & $-0.064$ & 60\% \\
$+\,$order $+$ phrasing            & 0.375 & $-0.105$ & 99\% \\
$+\,$order $+$ phrasing $+$ position & 0.374 & $-0.106$ & 100\% \\
\midrule
\multicolumn{4}{l}{\emph{(c) Non-semantic perturbation control}} \\
N0 baseline                        & 0.489 & ---      & --- \\
N1 whitespace-only                 & 0.485 & $-0.004$ & 4\% \\
N2 punctuation-only                & 0.481 & $-0.008$ & 8\% \\
N3 prefix-only                     & 0.476 & $-0.013$ & 13\% \\
N4 all three combined              & 0.475 & $-0.014$ & 13\% \\
Full PPA (reference)               & 0.374 & $-0.106$ & 100\% \\
\bottomrule
\end{tabular}%
}
\caption{PPA ablation on the 13-item development subset. (a) Each knob alone gives comparable effect size. (b) Position adds negligible value beyond order and phrasing. (c) Non-semantic perturbations capture only 13\% of PPA's TV reduction. Numerator and denominator of the ``\% of PPA'' column are both computed on this subset. Blocks (a)/(b) and (c) were run as separate batches with their own baseline calls (0.481 and 0.489); the reference row repeats the block-(b) PPA result.}
\label{tab:ppa-ablation}
\end{table}

\FloatBarrier
\section{PPA per-item results}
\label{app:ppa-per-item}

\paragraph{Per-cell breakdown ($2 \times 2$).} Table~\ref{tab:ppa-percell} reports baseline TV, PPA TV, and the paired Wilcoxon test within each of the four pre-registered (domain $\times$ entropy) cells.

\begin{table}[ht]
\centering
\small
\resizebox{\columnwidth}{!}{%
\begin{tabular}{lccccc}
\toprule
Cell & $n$ & Baseline TV & PPA TV & $\Delta$ & Wilcoxon $p$ \\
\midrule
FACTS $\times$ HI\_ENT  & 15 & 0.421 & 0.316 & $-0.104$ & 0.004 \\
FACTS $\times$ LO\_ENT  & 21 & 0.480 & 0.325 & $-0.155$ & $7.2 \times 10^{-4}$ \\
VALUES $\times$ HI\_ENT & 27 & 0.504 & 0.448 & $-0.056$ & 0.014 \\
VALUES $\times$ LO\_ENT & 37 & 0.430 & 0.337 & $-0.093$ & $2.6 \times 10^{-4}$ \\
\bottomrule
\end{tabular}%
}
\caption{Pre-registered $2 \times 2$ cell breakdown. Every cell improves; both factual cells confirm at $p < 0.01$.}
\label{tab:ppa-percell}
\end{table}

95\% bootstrap CIs on $\Delta$ (1000 resamples within each cell): FACTS-HI [$-0.16$, $-0.04$]; FACTS-LO [$-0.22$, $-0.09$]; VALUES-HI [$-0.09$, $-0.02$]; VALUES-LO [$-0.14$, $-0.05$].

\paragraph{Representative items.} Table~\ref{tab:ppa-per-item} lists four items from each cell, ordered by the size of PPA's improvement (near-duplicate battery items omitted), to illustrate where the gains concentrate.

\begin{table}[ht]
\centering
\small
\resizebox{\columnwidth}{!}{%
\begin{tabular}{llllrrr}
\toprule
Item & Wave & Domain & Entropy & Baseline TV & PPA TV & $\Delta$ \\
\midrule
NEIGHINTERA  & W32 & F & HI & 0.674 & 0.384 & $-0.290$ \\
WORRY2a      & W54 & F & HI & 0.493 & 0.212 & $-0.281$ \\
WORRY2d      & W54 & F & HI & 0.401 & 0.196 & $-0.206$ \\
TRACKCO1b    & W49 & F & HI & 0.564 & 0.414 & $-0.150$ \\
PP1          & W49 & F & LO & 0.682 & 0.194 & $-0.488$ \\
MADEUPSMCLICK& W45 & F & LO & 0.609 & 0.243 & $-0.366$ \\
FAMNEAR      & W32 & F & LO & 0.637 & 0.323 & $-0.314$ \\
NEIGHSAMEB   & W32 & F & LO & 0.466 & 0.166 & $-0.300$ \\
ECON5\_i     & W54 & V & HI & 0.654 & 0.381 & $-0.273$ \\
RACESURV5b   & W43 & V & HI & 0.678 & 0.448 & $-0.230$ \\
RACESURV47d  & W43 & V & HI & 0.568 & 0.416 & $-0.152$ \\
LEGALIMMIGAMT& W92 & V & HI & 0.464 & 0.314 & $-0.150$ \\
RESTRICTWHO  & W45 & V & LO & 0.874 & 0.255 & $-0.619$ \\
GAP21Q15\_f  & W82 & V & LO & 0.699 & 0.369 & $-0.330$ \\
RACESURV41   & W43 & V & LO & 0.563 & 0.253 & $-0.310$ \\
RACESURV5a   & W43 & V & LO & 0.357 & 0.074 & $-0.283$ \\
\bottomrule
\end{tabular}%
}
\caption{Representative items, ordered by PPA improvement. F = factual; V = values; HI/LO = entropy bin.}
\label{tab:ppa-per-item}
\end{table}

\FloatBarrier
\section{Baseline sweeps}
\label{app:sweeps}

Decoding-parameter sweeps on the same 13-item pre-registered development subset used in Appendix~\ref{app:ablation}, 100 personas per item.

\begin{table}[ht]
\centering
\small
\begin{tabular}{ccc}
\toprule
$T$ & Mean TV ($n=13$) & $\Delta$ vs.\ $T = 1.0$ \\
\midrule
0.7 & 0.490 & $+0.011$ \\
1.0 & 0.479 & 0 (baseline) \\
1.3 & 0.471 & $-0.008$ \\
1.5 & 0.463 & $-0.016$ \\
\bottomrule
\end{tabular}%
\caption{Temperature sweep. Best $T = 1.5$ captures $\approx 15\%$ of PPA's gain.}
\label{tab:temp-sweep}
\end{table}

\begin{table}[ht]
\centering
\small
\begin{tabular}{ccc}
\toprule
$p$ & Mean TV ($n=13$) & $\Delta$ vs.\ $p = 1.0$ \\
\midrule
0.5 & 0.490 & $+0.001$ \\
0.7 & 0.493 & $+0.004$ \\
0.9 & 0.482 & $-0.007$ \\
1.0 & 0.489 & 0 (baseline) \\
\bottomrule
\end{tabular}%
\caption{Top-$p$ sweep. Best $p = 0.9$ captures $\approx 7\%$ of PPA's gain.}
\label{tab:topp-sweep}
\end{table}

Neither sweep approaches PPA's effect size; the dominant factor in PPA's improvement is the semantic perturbation, not the decoding distribution.

\FloatBarrier
\section{PPA mechanism: non-semantic controls and temperature analogue}
\label{app:ppa-mechanism}

This appendix details the two follow-up tests reported in \S\ref{sec:entropy-mechanism}. The aim is to ask whether the entropy-correlation result reflects the mechanism we claim --- that PPA helps by raising per-call entropy --- or some alternative such as ``any prompt variation helps'' or ``the gain is specific to semantic perturbation.''

\paragraph{Non-semantic controls.} On the same 20-item subset, we compare PPA's per-call entropy increase against four non-semantic perturbation conditions matched to PPA in structure but not in content: whitespace variation (N1), punctuation variation (N2), prefix variation (N3), and all three combined (N4). Each non-semantic condition perturbs surface form across 50 calls per (persona, item) cell without changing option ordering or question phrasing.

\begin{table}[ht]
\centering
\small
\begin{tabular}{lrrr}
\toprule
Condition & Mean $H$ (nats) & $\Delta H$ vs.\ std & $\Delta\text{TV}$ \\
\midrule
Standard Argyle    & 0.145 & ---       & ---       \\
PPA                & 0.562 & $+0.417$  & $+0.112$  \\
N1 whitespace      & 0.146 & $+0.000$  & $-0.001$  \\
N2 punctuation     & 0.138 & $-0.007$  & $+0.001$  \\
N3 prefix          & 0.157 & $+0.011$  & $+0.002$  \\
N4 all three       & 0.149 & $+0.003$  & $-0.002$  \\
\bottomrule
\end{tabular}
\caption{Per-cell entropy and TV-to-Pew on the 20-item subset under PPA and four non-semantic perturbation conditions. PPA raises per-call entropy by 0.417 nats; the four non-semantic conditions average $\Delta H = 0.002$.}
\label{tab:ppa-nonsemantic}
\end{table}

The four non-semantic conditions raise per-call entropy by essentially zero on average ($\Delta H = 0.002$), two orders of magnitude smaller than PPA's $\Delta H = 0.417$, and yield TV changes between $-0.002$ and $+0.002$, indistinguishable from zero against PPA's $+0.112$. The TV improvement tracks the entropy increase, not the presence of prompt variation.

\paragraph{Temperature analogue.} We compare PPA's $\Delta\text{TV}/\Delta H$ slope against an independent intervention that also raises per-call entropy: decoding temperature. Re-running standard Argyle at $T = 1.5$ on the same 20-item subset (50 calls per cell) raises per-call entropy by $\Delta H = 0.069$ and reduces TV by $0.015$, a slope of 0.21 against PPA's 0.27 (Table~\ref{tab:entropy-mechanism}).

Semantic prompt perturbation and decoding-temperature increase have very different mechanics but produce a similar TV improvement per nat of entropy. Combined with the non-semantic null result, this shows that the amount of entropy added, not the specific surface knob used to add it, governs the aggregate improvement. This is a statement about magnitude, not about structure: both interventions draw from the model's own persona-conditioned distribution rather than from a flat prior, so neither behaves like uniform noise. Section~\ref{sec:ppa-structure} shows directly that PPA's added variation is structured, which is what separates it from noise of the same magnitude.

\paragraph{Bootstrap stability.} The entropy-correlation regression of \S\ref{sec:entropy-mechanism} uses $N = 20$ items. We assess the robustness of the $\beta_{\Delta H}$ coefficient by resampling with replacement 1000 times. The coefficient is positive in 1000/1000 resamples (mean $0.270$, 95\% bootstrap CI $[0.176, 0.373]$). 

\FloatBarrier
\section{PPA subgroup structure}
\label{app:ppa-subgroup}

We test whether PPA's added variation is structured by political lean rather than uniform. Each persona carries Pew-sampled demographic attributes including a political lean (Democrat or Democrat-leaning vs.\ Republican or Republican-leaning). We use the 50 items with sufficient respondents in both groups. For each item we group answers by lean and run a chi-square test of independence.

\begin{table}[ht]
\centering
\small
\resizebox{\columnwidth}{!}{%
\begin{tabular}{lcc}
\toprule
 & Standard Argyle & PPA \\
\midrule
Items with a lean difference ($\chi^2$) & 27/50 & \textbf{33/50} \\
Items with single-answer collapse       & 6/50  & \textbf{2/50} \\
\bottomrule
\end{tabular}%
}
\caption{Per-item chi-square test of answer distribution by political lean. Under PPA, more items show a significant Democrat vs.\ Republican difference and fewer collapse to a single answer. Uniform noise would reduce, not increase, the number of items with a subgroup difference.}
\label{tab:ppa-subgroup}
\end{table}

\paragraph{Gap alignment with Pew.} For each item we compute the Democrat-minus-Republican gap in Pew and the corresponding gap between Democrat- and Republican-leaning personas in the model, and compare the two across the 25 items on which both party subgroups are measurable on both sides (at least 30 Pew respondents and at least 10 persona responses per party).

\begin{table}[ht]
\centering
\small
\begin{tabular}{lcc}
\toprule
Metric & Standard Argyle & PPA \\
\midrule
Correlation with Pew gap ($r$) & 0.573 & \textbf{0.678} \\
Sign agreement with Pew gap    & 64\%  & \textbf{76\%} \\
\bottomrule
\end{tabular}
\caption{Alignment between the model's per-item Democrat--Republican gap and the Pew gap. PPA moves the model's subgroup gaps closer to the human ones.}
\label{tab:ppa-gap}
\end{table}

\paragraph{PPA converges toward the model's own described distribution.} If PPA merely spread mass, its pooled estimate would drift toward uniform. Instead it moves toward the distribution the same model produces under the describe pathway. For the two open models of Table~\ref{tab:openmodel-main} whose describe output is a JSON distribution, Table~\ref{tab:ppa-convergence} reports, per item, the TV between the pooled persona answers (standard Argyle or PPA) and (i) the model's own described distribution and (ii) the uniform distribution.

\begin{table}[ht]
\centering
\small
\resizebox{\columnwidth}{!}{%
\begin{tabular}{lcccc}
\toprule
Model & $n$ & TV(Argyle, describe) & TV(PPA, describe) & TV(PPA, uniform) \\
\midrule
Qwen3-8B            & 100 & 0.507 & \textbf{0.291} & 0.360 \\
Qwen2.5-7B-Instruct & 64  & 0.556 & \textbf{0.375} & 0.448 \\
\bottomrule
\end{tabular}%
}
\caption{PPA moves the pooled estimate toward the model's own described distribution. Argyle-to-describe vs.\ PPA-to-describe, paired Wilcoxon: $p = 2 \times 10^{-17}$ (Qwen3-8B) and $9 \times 10^{-11}$ (Qwen2.5-7B-Instruct); PPA is closer to the described distribution than to uniform on 73 of 100 and 49 of 64 items ($p = 8 \times 10^{-9}$ and $1 \times 10^{-6}$). $n$ = items whose describe output is a JSON distribution (Qwen2.5-7B-Instruct's remaining 18 parsed items were prose).}
\label{tab:ppa-convergence}
\end{table}

\paragraph{PPA and demographic matching are independent fixes.} A separate concern is whether PPA still helps once personas are demographically matched to each item's respondents (\S\ref{sec:silicon}). For each of the 100 items we draw 100 personas matched to the item's Pew respondents and compare matched sampling against matched sampling with PPA, 100 calls per item per arm, across four models. PPA improves matched sampling on every model (Table~\ref{tab:matched-ppa}). Matching changes which personas are asked; PPA changes how each persona answers. The two are independent, and better matching does not remove the need for PPA.

\begin{table}[ht]
\centering
\small
\resizebox{\columnwidth}{!}{%
\begin{tabular}{lccc}
\toprule
Model & Matched TV & Matched $+$ PPA TV & $p$ \\
\midrule
GPT-4o                & 0.482 & \textbf{0.394} & $2 \times 10^{-7}$ \\
Qwen2.5-7B-Instruct   & 0.543 & \textbf{0.346} & $3 \times 10^{-15}$ \\
Llama-3.1-8B-Instruct & 0.426 & \textbf{0.285} & $1 \times 10^{-14}$ \\
Qwen3-8B              & 0.491 & \textbf{0.315} & $2 \times 10^{-14}$ \\
\bottomrule
\end{tabular}%
}
\caption{Matched Argyle vs.\ matched Argyle with PPA, 100 items, 100 calls per item per arm. PPA reduces TV on every model, so its benefit does not overlap with demographic matching. The GPT-4o matched arm is an independent re-run of the Appendix~\ref{app:matched-argyle} matched pipeline.}
\label{tab:matched-ppa}
\end{table}

\FloatBarrier
\section{Non-Likert items}
\label{app:non-likert}

We select 24 Pew items with 2, 3, or 4 answer options (8 per option count), each with at least 200 respondents, and run standard Argyle, describe, and PPA with 100 persona calls per item per pipeline on three open models (Table~\ref{tab:non-likert}).

\begin{table}[ht]
\centering
\small
\resizebox{\columnwidth}{!}{%
\begin{tabular}{llccc}
\toprule
Model & $k$ & Argyle & PPA & Describe \\
\midrule
Llama-3.1-8B-Instruct & 2 & 0.177 & 0.158 & 0.139 \\
 & 3 & 0.289 & 0.270 & 0.229 \\
 & 4 & 0.296 & 0.266 & 0.182 \\
 & all & 0.254 & 0.232 & \textbf{0.184} \\
Qwen2.5-7B-Instruct & 2 & 0.150 & 0.222 & 0.091 \\
 & 3 & 0.378 & 0.277 & 0.268 \\
 & 4 & 0.368 & 0.297 & 0.236 \\
 & all & 0.299 & 0.265 & \textbf{0.199} \\
Qwen3-8B & 2 & 0.198 & 0.204 & 0.182 \\
 & 3 & 0.450 & 0.170 & 0.196 \\
 & 4 & 0.439 & 0.356 & 0.228 \\
 & all & 0.362 & 0.243 & \textbf{0.202} \\
\bottomrule
\end{tabular}%
}
\caption{Non-Likert benchmark: mean TV-to-Pew by model and option count $k$ (8 items per $k$; 100 calls per item per pipeline). Describe vs.\ Argyle, paired two-sided Wilcoxon over 24 items: $p = 0.025$, $0.006$, $0.001$ (top to bottom); Argyle vs.\ PPA on Qwen3-8B: $p = 0.012$.}
\label{tab:non-likert}
\end{table}

\FloatBarrier
\section{Open-ended generation}
\label{app:openended}

Five open-ended survey tasks (most important problem, main news source, top policy priority, occupation, U.S. city of residence), four instruction-tuned models, 200 calls per condition at $T = 1.0$. Three conditions per task: the fixed prompt; six non-semantic whitespace/punctuation transforms of that prompt, one drawn per call; and ten semantic paraphrases, one drawn per call (all paraphrases are listed verbatim in the attached supplementary repository). We measure per-call answer entropy after normalizing surface form (case, punctuation, articles). Table~\ref{tab:openended} reports seed 13; seed 47 reproduces every cell's direction (mean paraphrase gain $+1.01$, non-semantic $+0.02$, Spearman $\rho = -0.76$).

\begin{table}[ht]
\centering
\small
\resizebox{\columnwidth}{!}{%
\begin{tabular}{llrrrr}
\toprule
Model & Task & Fixed & Non-sem. & Paraphrase & $\Delta$ \\
\midrule
Llama-3.1-8B-Instruct & important problem & 4.81 & 4.79 & 4.61 & $-0.20$ \\
 & news source       & 3.44 & 3.48 & 4.20 & $+0.76$ \\
 & policy priority   & 4.09 & 3.99 & 4.86 & $+0.76$ \\
 & occupation        & 3.82 & 3.96 & 4.34 & $+0.53$ \\
 & city of residence & 3.35 & 3.34 & 3.72 & $+0.37$ \\
\midrule
Qwen2.5-7B-Instruct & important problem & 1.26 & 1.36 & 2.92 & $+1.66$ \\
 & news source       & 1.65 & 1.70 & 3.38 & $+1.73$ \\
 & policy priority   & 2.08 & 2.09 & 4.15 & $+2.08$ \\
 & occupation        & 0.73 & 0.88 & 1.91 & $+1.18$ \\
 & city of residence & 1.84 & 1.51 & 1.89 & $+0.05$ \\
\midrule
Qwen3-8B & important problem & 1.95 & 2.01 & 2.72 & $+0.77$ \\
 & news source       & 1.18 & 1.18 & 3.19 & $+2.01$ \\
 & policy priority   & 1.68 & 1.73 & 3.48 & $+1.80$ \\
 & occupation        & 1.72 & 1.96 & 3.18 & $+1.45$ \\
 & city of residence & 0.38 & 0.50 & 2.37 & $+1.99$ \\
\midrule
OLMo-2-7B-Instruct & important problem & 2.04 & 2.18 & 3.10 & $+1.06$ \\
 & news source       & 4.65 & 4.53 & 4.05 & $-0.60$ \\
 & policy priority   & 2.71 & 2.63 & 3.83 & $+1.12$ \\
 & occupation        & 1.83 & 1.72 & 2.80 & $+0.97$ \\
 & city of residence & 3.27 & 3.20 & 3.45 & $+0.18$ \\
\bottomrule
\end{tabular}%
}
\caption{Per-call answer entropy (nats) under the fixed prompt, non-semantic surface variation, and semantic paraphrase; $\Delta$ = paraphrase $-$ fixed. 200 calls per cell, $T = 1.0$, seed 13. Mean $\Delta$: paraphrase $+0.98$ (16 of 20 cells $\geq 0.2$), non-semantic $+0.01$. The gain tracks how collapsed the fixed prompt is (Spearman $\rho = -0.75$, $p = 1.2 \times 10^{-4}$) and turns negative only on the two cells already above 4.6 nats.}
\label{tab:openended}
\end{table}

\FloatBarrier
\section{V8 internal structure}
\label{app:v8-structure}

To check whether V8 is producing genuinely independent draws or constructing a parametric description, we examine three properties of its $N = 20$ outputs across the 10 lists per task (\S\ref{sec:failure-knowsdoes}). Under independent sampling (Hypothesis A), within-call lag-1 autocorrelation should be near zero, cross-call sequence-rank correlation should be at chance, and run-length statistics should match a binomial null. Under construction (Hypothesis B), at least one of these signatures should depart sharply from those predictions.

\begin{table}[ht]
\centering
\small
\resizebox{\columnwidth}{!}{%
\begin{tabular}{lrrrc}
\toprule
Task & Within-call lag-1 autocorr & Cross-call seq-rank $\rho$ & Run-length ratio & Hypothesis \\
\midrule
uniform 1--5 & $-0.221$ & 1.000 & 0.820 & B (constructs) \\
fair coin     & $-0.494$ & 1.000 & 0.665 & B (constructs) \\
skewed binary & $-0.039$ & 1.000 & 1.213 & B (constructs) \\
bimodal 5-way & $+0.282$ & 0.995 & 2.596 & B (constructs) \\
skewed 5-way  & $+0.550$ & 0.966 & 2.436 & B (constructs) \\
\bottomrule
\end{tabular}%
}
\caption{V8 internal structure. Every task assigns to construction (Hypothesis B) by cross-call sequence-rank correlation $\geq 0.96$.}
\label{tab:v8-structure}
\end{table}

The cross-call sequence-rank correlation is the cleanest diagnostic: across the 10 independently sampled lists, the ordering of the 20 items inside each call agrees with the ordering in every other call at $\rho \geq 0.96$. Independent draws cannot produce that. Within-call autocorrelation also departs from the independence null in opposite directions on different tasks (strongly negative on the fair coin, where the model alternates H-T-H-T; strongly positive on the skewed 5-way, where the model groups runs of repeated values that match the target proportions). Both signatures point to V8 constructing a single fixed sequence each call rather than drawing 20 independent samples.

\FloatBarrier
\section{Cross-family replication}
\label{app:cross-family}

To address the single-model scope of the main paper, we replicate the structural claims on five additional instruction-tuned model families; $N=200$ calls per cell (V8: 10 lists of 20) at $T=1.0$.

\paragraph{Models.}
\texttt{openai/gpt-5.4}, \texttt{anthropic/claude-sonnet-4.5}, \texttt{meta-llama/llama-3.3-70b-instruct}, \texttt{google/gemini-3-flash-preview}, \texttt{deepseek/deepseek-chat} (V3.2).

\paragraph{RNG probe.} Every instruction-tuned model rejects uniformity on the RNG probe; the favored value varies across families.

\begin{table}[ht]
\centering
\small
\resizebox{\columnwidth}{!}{%
\begin{tabular}{lcccc}
\toprule
Model & int 1--100 fav. & $P(\text{mode})$ & coverage & KS $p$ on float $[0,1]$ \\
\midrule
\texttt{gpt-4o} (main paper) & 42 & 0.775 & 15/100 & $5 \times 10^{-25}$ \\
\texttt{gpt-5.4}        & 73 & 0.110 & 59/100 & $< 10^{-24}$ \\
\texttt{claude-sonnet-4.5} & 47 & 1.000 & 1/100  & $< 10^{-30}$ \\
\texttt{llama-3.3-70b}  & 53 & 0.890 & 4/100  & $< 10^{-30}$ \\
\texttt{gemini-3-flash} & 42 & 0.910 & 8/100  & $< 10^{-29}$ \\
\texttt{deepseek-chat}  & 1  & 0.330 & 18/100 & $< 10^{-30}$ \\
\bottomrule
\end{tabular}%
}
\caption{Cross-family RNG probe. Every instruction-tuned model concentrates on a favorite value.}
\label{tab:cross-family-rng}
\end{table}

\paragraph{Seven-target panel.} Table~\ref{tab:cross-family-seven} reports $P(\text{mode})$ on each of the seven targets from the main paper across the five additional model families. Every model concentrates at $P(\text{mode}) \geq 0.46$ on at least one non-point target; Claude-Sonnet-4.5 hits $P(\text{mode}) = 1.0$ on six of the seven.

\begin{table}[ht]
\centering
\small
\resizebox{\columnwidth}{!}{%
\begin{tabular}{lccccc}
\toprule
Target & gpt-5.4 & claude-4.5 & llama-70b & gemini-3 & deepseek \\
\midrule
point 5            & 1.000 & 1.000 & 1.000 & 1.000 & 1.000 \\
mixture 5/4        & 0.860 & 1.000 & 1.000 & 1.000 & 0.887 \\
skewed binary      & 0.770 & 1.000 & 0.830 & 0.990 & 0.990 \\
bimodal 1/5        & 0.650 & 0.950 & 0.945 & 0.775 & 0.792 \\
three-way          & 0.945 & 1.000 & 0.985 & 0.663 & 0.720 \\
skewed 5-way       & 0.500 & 1.000 & 0.985 & 0.610 & 0.870 \\
uniform 1--5       & 0.460 & 1.000 & 0.940 & 0.610 & 0.860 \\
\midrule
$\geq 0.90$ count  & 2/7   & 7/7   & 6/7   & 3/7   & 2/7 \\
\bottomrule
\end{tabular}%
}
\caption{Cross-family seven-target panel: $P(\text{mode})$ on each target.}
\label{tab:cross-family-seven}
\end{table}

\paragraph{KNOWS/DOES dichotomy.} Table~\ref{tab:cross-family-seven} uses the seven targets of Table~\ref{tab:seven-targets}; Tables~\ref{tab:cross-family-v8} and~\ref{tab:cross-family-v0} use the five-target panel of Table~\ref{tab:base-instruct}, in separate runs.

\begin{table}[ht]
\centering
\small
\resizebox{\columnwidth}{!}{%
\begin{tabular}{lccccc}
\toprule
& \multicolumn{5}{c}{V8 TV-to-target} \\
\cmidrule(lr){2-6}
Task & gpt-5.4 & claude-4.5 & llama-70b & gemini-3 & deepseek \\
\midrule
uniform 1--5   & 0.000 & 0.000 & 0.008 & 0.000 & 0.000 \\
fair coin       & 0.000 & 0.050 & 0.000 & 0.000 & 0.000 \\
skewed binary   & 0.000 & 0.000 & 0.000 & 0.000 & 0.000 \\
bimodal 5-way   & 0.000 & 0.000 & 0.015 & 0.000 & 0.000 \\
skewed 5-way    & 0.000 & 0.000 & 0.000 & 0.000 & 0.000 \\
\bottomrule
\end{tabular}%
}
\caption{V8 KNOWS TV-to-target across families. V8 succeeds universally: max TV = 0.050.}
\label{tab:cross-family-v8}
\end{table}

\begin{table}[ht]
\centering
\small
\resizebox{\columnwidth}{!}{%
\begin{tabular}{lccccc}
\toprule
Task & gpt-5.4 & claude-4.5 & llama-70b & gemini-3 & deepseek \\
\midrule
uniform 1--5   & 0.295 & 0.800 & 0.735 & 0.600 & 0.630 \\
fair coin       & 0.195 & 0.500 & 0.269 & 0.055 & 0.455 \\
skewed binary   & 0.130 & 0.300 & 0.140 & 0.295 & 0.300 \\
bimodal 5-way   & 0.500 & 0.240 & 0.600 & 0.320 & 0.220 \\
skewed 5-way    & 0.350 & 0.500 & 0.630 & 0.490 & 0.365 \\
\bottomrule
\end{tabular}%
}
\caption{V0 (plain DOES) baseline TV per (model, task). V0 fails on 21 of 25 cells (TV $> 0.20$).}
\label{tab:cross-family-v0}
\end{table}

\paragraph{Summary.} The structural claims generalize qualitatively across five instruction-tuned families: every model rejects uniformity on the RNG probe (five distinct favorites across the six models), every model exhibits categorical or strongly attenuated collapse on most non-uniform targets, and V8 succeeds across all 25 (model, task) cells. Combined with the base-vs-instruct comparison of \S\ref{sec:failure-alignment}, this is consistent with instruction-targeted post-training as the proximate cause of the collapse.

\FloatBarrier
\end{document}